\documentclass{article}
\usepackage{iclr2021_conference,times}

\usepackage{wrapfig}
\usepackage[bookmarks=true]{hyperref}
\usepackage{tikz}
\usepackage{graphicx}
\usepackage{relsize}
\usepackage{adjustbox}
\usepackage{listings}
\usepackage{textcomp}
\usepackage{color}
\usepackage{mathtools}
\usepackage{amsmath}
\usepackage{flushend}
\usepackage{pifont}

\usepackage{epsfig,graphicx,amsmath,amsfonts}
\usepackage{import}
\usepackage{transparent}
\usepackage{upgreek}

\definecolor{mintedbgcolor}{rgb}{0.85,0.85,0.85}
\usepackage{xcolor}
\usepackage{tcolorbox}

\newcommand{\cmark}{\textcolor[rgb]{0,0.6,0}{\ding{51}}}%
\newcommand{\xmark}{\textcolor[rgb]{0.6,0,0}{\ding{55}}}%

\definecolor{codegreen}{rgb}{0,0.4,0}
\definecolor{codegray}{rgb}{1.0,0.5,0.5}
\definecolor{codepurple}{rgb}{0.58,0,0}
\definecolor{codebackcolour}{rgb}{0.95,0.95,0.92}

\lstdefinestyle{mystyle}{
    backgroundcolor=\color{codebackcolour},   
    commentstyle=\bf\color{codegreen},
    keywordstyle=\color{magenta},
    stringstyle=\color{codepurple},
    basicstyle=\ttfamily\tiny,
    breakatwhitespace=false,         
    breaklines=true,                 
    captionpos=b,                    
    keepspaces=true,                 
    numbers=none,                    
    showspaces=false,                
    showstringspaces=false,
    showtabs=false,                  
    tabsize=2
}

\usepackage[boxruled,linesnumbered]{algorithm2e}
\usepackage[normalem]{ulem}
\usepackage{multirow}
\usepackage[utf8]{inputenc}
\usepackage[TS1,T1]{fontenc}
\usepackage{array, booktabs}
\usepackage{xcolor}

\usepackage{caption}

\title{\bf
NViSII: A Scriptable Tool for \\
Photorealistic Image Generation
}

\author{%
Nathan Morrical$^{1,2}$,
Jonathan Tremblay$^1$,
Yunzhi Lin$^{1,3}$,\\
\bf{Stephen Tyree}$^1$,
\bf{Stan Birchfield}$^1$,
\bf{Valerio Pascucci}$^2$, 
\bf{Ingo Wald}$^1$\\
\vspace{2mm}
$^1$NVIDIA,
$^2$University of Utah,
$^3$Georgia Institute of Technology
}

\iclrfinalcopy
\begin{document}

\makeatletter
\newcommand{\definetrim}[2]{%
    \define@key{Gin}{#1}[]{\setkeys{Gin}{trim=#2,clip}}%
}
\makeatother
\definetrim{custom_trim}{800px 200px 800px 200px}

\begin{center}
    \maketitle
    \vspace{-1.0em}
    \centering
    \resizebox{0.98\textwidth}{!}{%
        \setlength{\fboxrule}{0pt}%
        \setlength{\fboxsep}{0pt}%
        \begin{tikzpicture}%
        \node[anchor=south west,inner sep=0] (image) at (0,0) {\includegraphics[trim=0px 100px 0px 50px,clip]{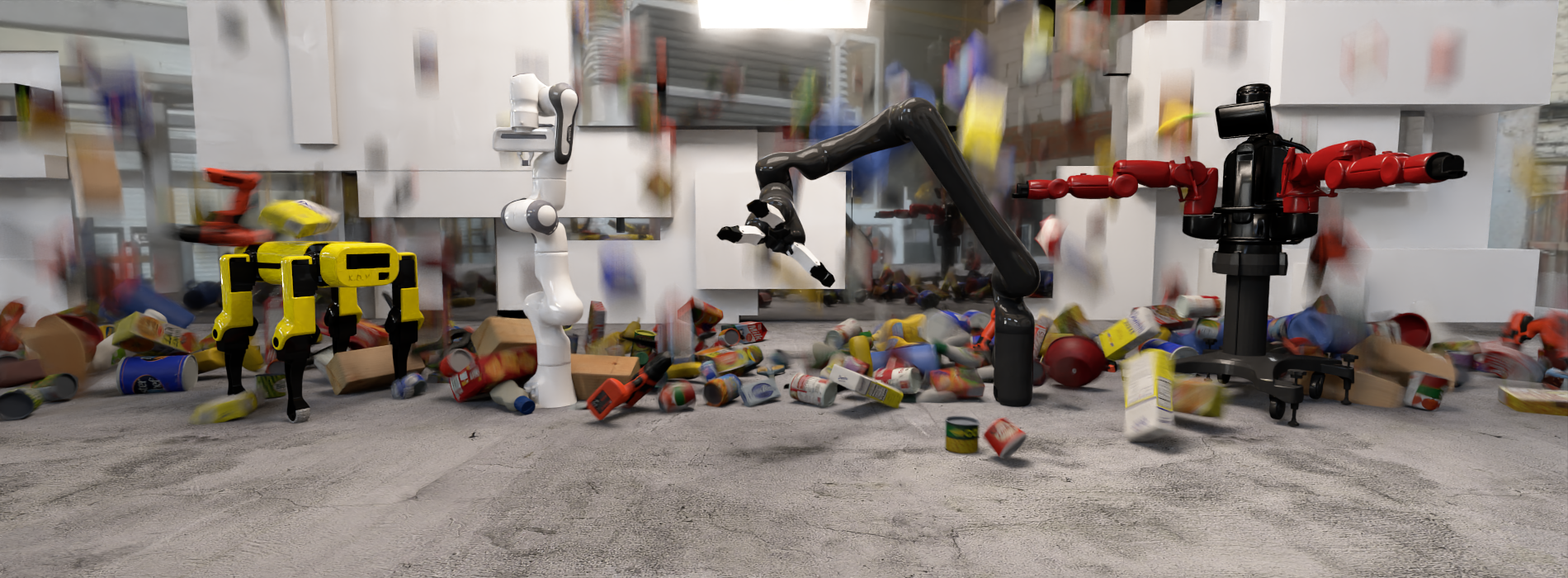}};
    	\begin{scope}[x={(image.south east)},y={(image.north west)}] %
    	\end{scope}%
        \end{tikzpicture}%
     }
     \resizebox{.98\textwidth}{!}{%
        \includegraphics[custom_trim]{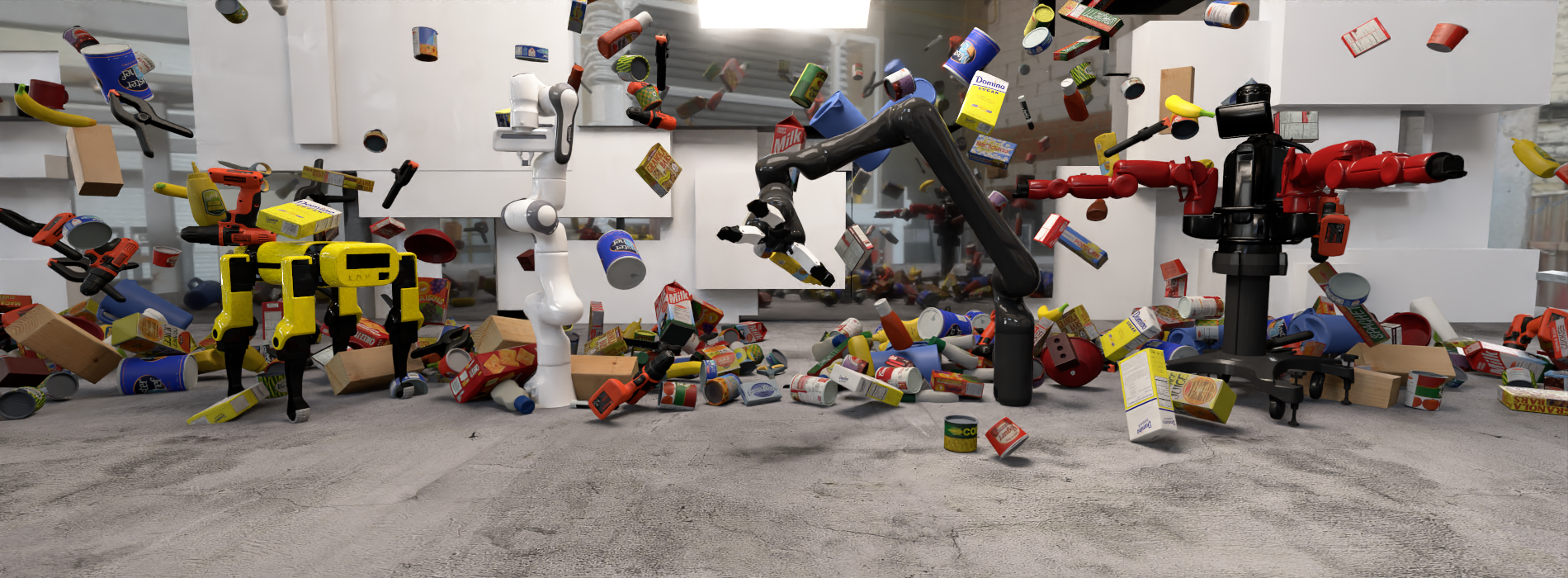}%
        \hspace{0.05pt}%
        \includegraphics[custom_trim]{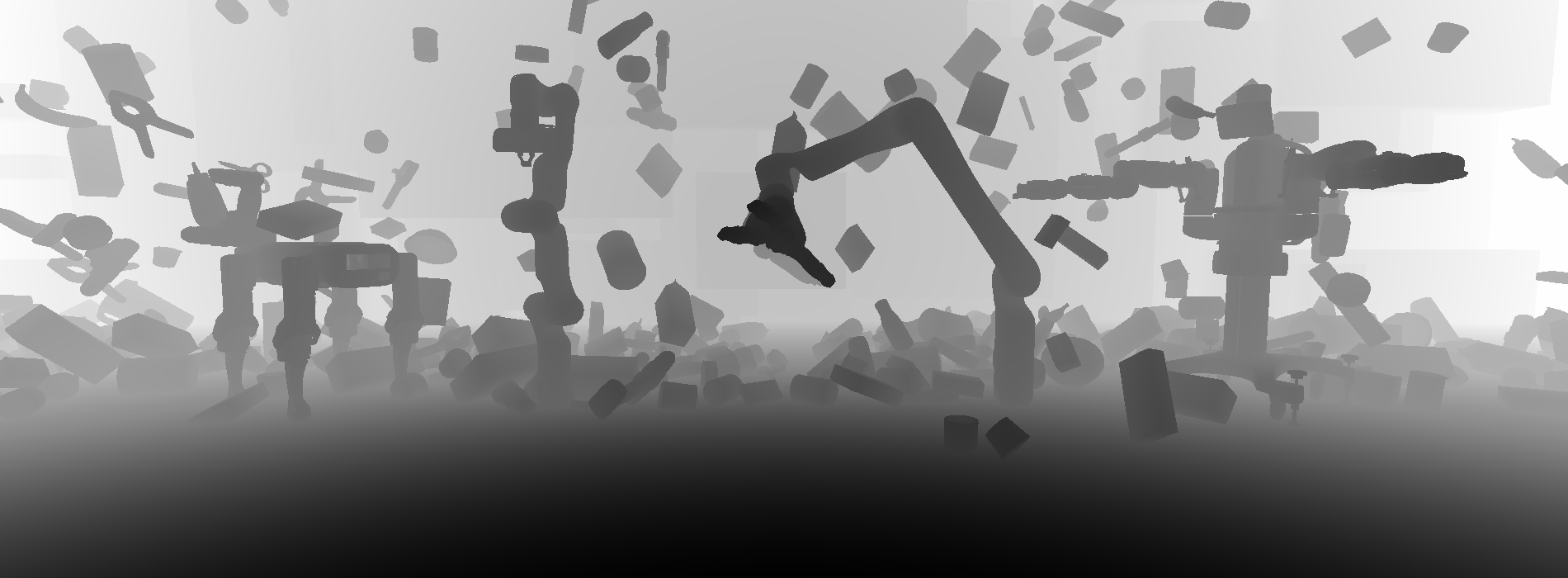}%
        \hspace{0.05pt}%
    	\includegraphics[custom_trim]{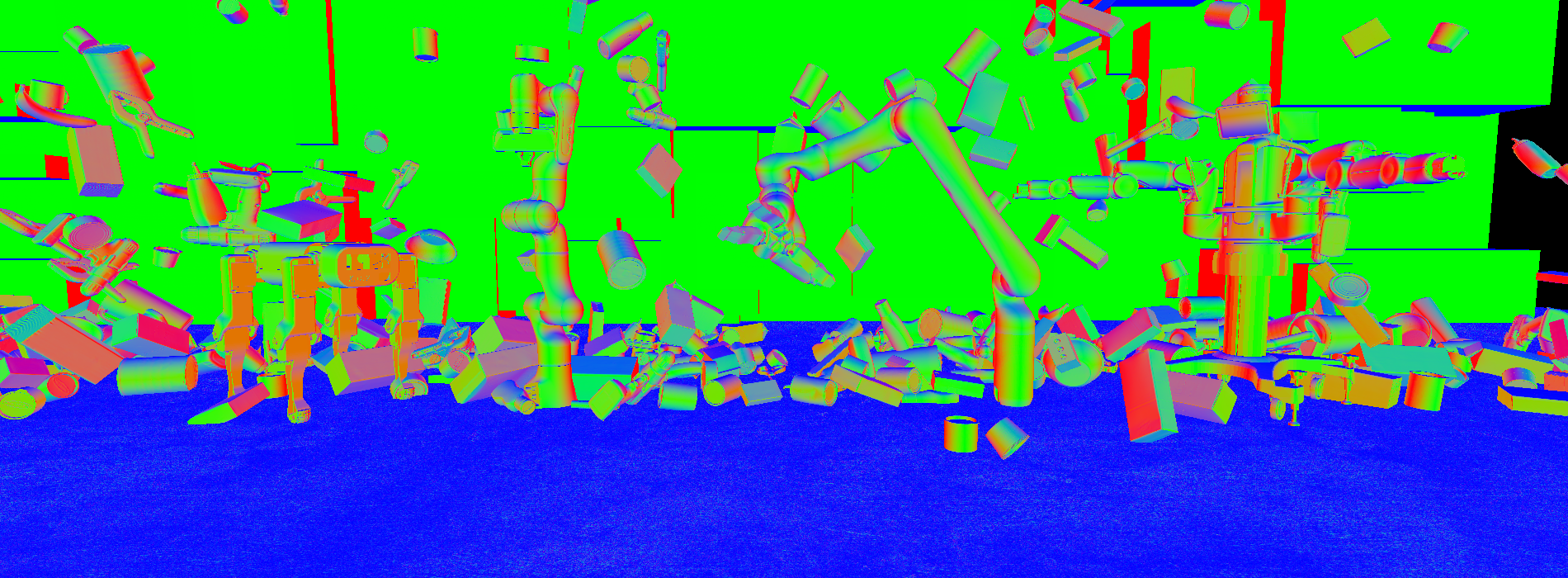}%
        \hspace{0.05pt}%
        \includegraphics[custom_trim]{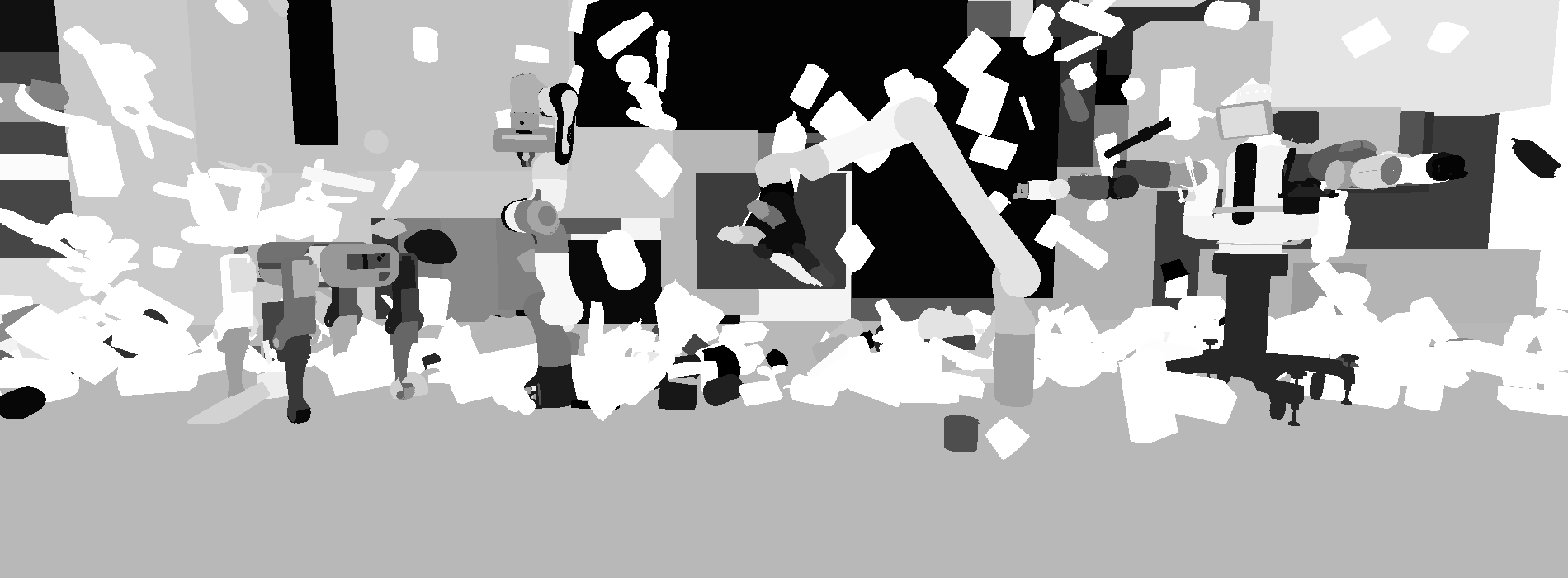}%
        \hspace{0.05pt}%
        \includegraphics[custom_trim]{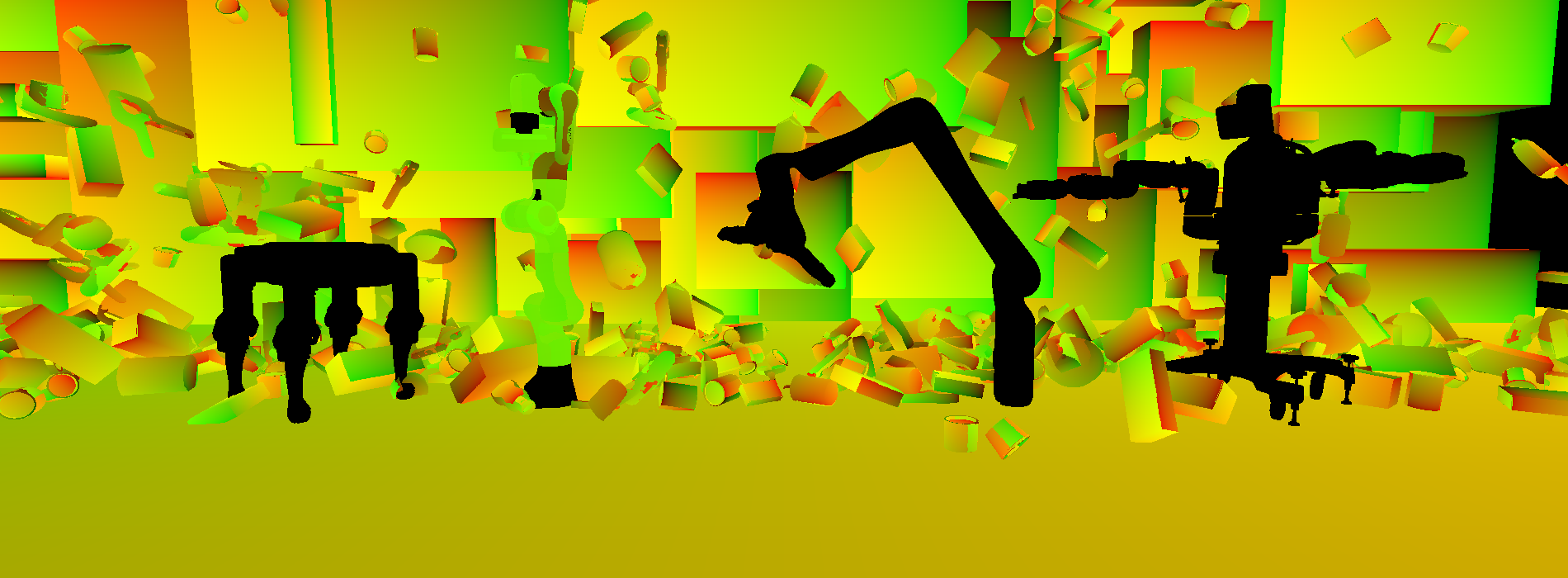}%
        \hspace{0.05pt}%
        \includegraphics[custom_trim]{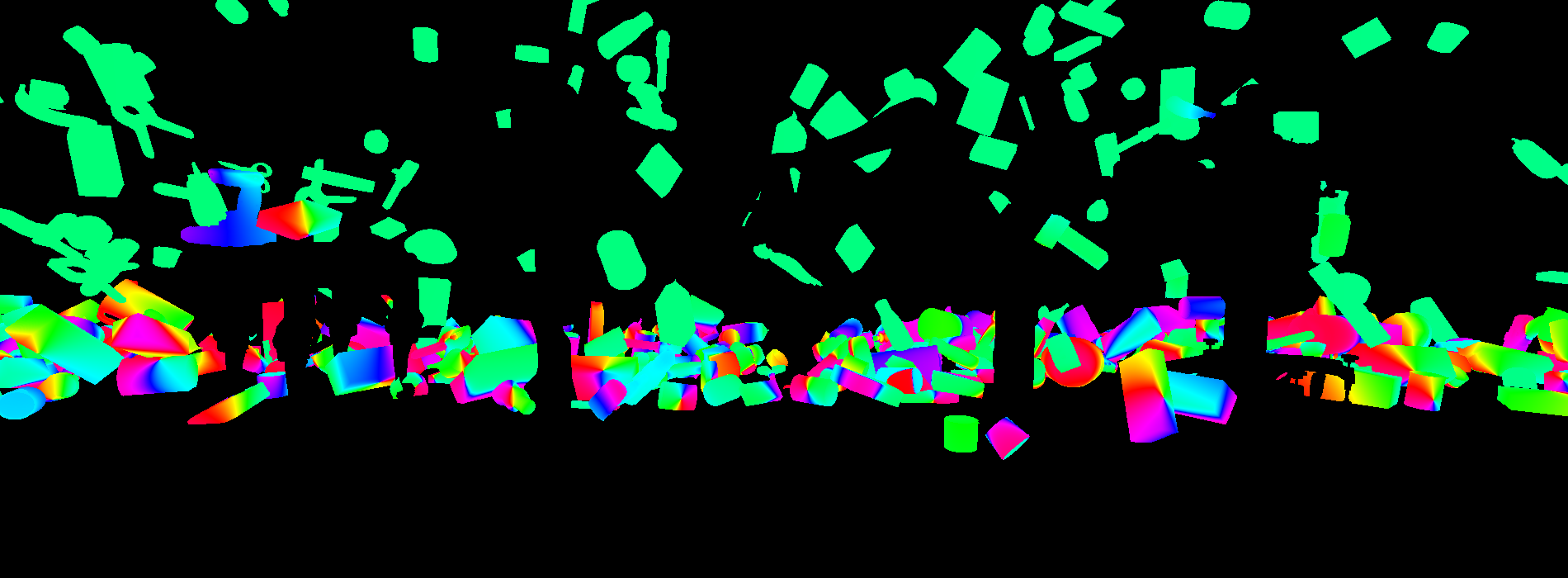}%
      }
    \captionof{figure}{Our scriptable tool leverages hardware-accelerated path tracing to generate photorealistic images. Annotations are shown along the bottom row, from left to right:
    image with no motion blur, depth, surface normals, segmentation labels, texture coordinates, and optical flow.}
    \label{fig:teaser}
    \vspace{-0.0em}
\end{center}

\begin{abstract}
We present a Python-based renderer built on NVIDIA's OptiX ray tracing engine and the OptiX AI denoiser, designed to generate high-quality synthetic images for research in computer vision and deep learning.  
Our tool enables the description and manipulation of complex dynamic 3D scenes containing object meshes, materials, textures, lighting, 
volumetric data (\emph{e.g.}, smoke),
and backgrounds. 
Metadata, such as 2D/3D bounding boxes, segmentation masks, depth maps, normal maps, material properties, and optical flow vectors, can also be generated. 
In this work, we discuss design goals, architecture, and performance. 
We demonstrate the use of data generated by path tracing 
for training an object detector and pose estimator, showing improved performance in sim-to-real transfer in situations that are difficult for traditional raster-based renderers. 
We offer this tool as an easy-to-use, performant, high-quality renderer for advancing research in synthetic data generation and deep learning.
\end{abstract}

\section{Introduction}

For many computer vision tasks, it is challenging or even impossible to obtain labeled real-world images for use in training deep neural networks.
For example, labeled ground truth of rare events like vehicle collisions, or of dense high-dimensional values like optical flow vectors, are not easy to obtain. 
To overcome these limitations, researchers have explored synthetic data for a variety of applications: 
object pose estimation \citep{tremblay2018arx:fat,denninger2019blenderproc}, 
depth estimation of transparent objects \citep{sajjan2019cleargrasp},
scene segmentation \citep{handa2015arx:sn,hypersim},
optical flow \citep{dosovitskiy2015iccv:flownet},
autonomous vehicles \citep{ros2016cvpr:syn,prakash2018structured},
robotic control \citep{tobin2017iros:dr},
path planning and reasoning in 3D scenes \citep{kolve2017ai2,xia2020interactive}, 
and so forth.

To generate such datasets, a variety of tools have been developed, including
AI2-Thor \citep{kolve2017ai2},
iGibson \citep{xia2020interactive},
NDDS \citep{to2018ndds},
Unity3D \citep{unity2020perception},
Sapien \citep{xiang2020sapien},
and BlenderProc \citep{denninger2019blenderproc},
as shown in Table~\ref{tab:relatedwork}.
Although AI2-Thor and iGibson come with powerful Python APIs, both of them are based on classic raster scanning.
On the other hand, the more recent tools capable of photorealistic imagery via path tracing (such as Unity3D, Sapien, and BlenderProc) come with limited scriptable 
interfaces, such as runtime scene editing, limited material and light expression, {\em etc}.
To overcome this limitation, in this work we introduce 
NVISII: the NVIDIA Scene Imaging Interface,
a scriptable tool for path-traced image generation. 
With our tool, users can construct and manipulate complex dynamic 3D scenes containing object meshes, materials, textures, lighting, volumetric data (\emph{e.g.}, smoke), and cameras---all potentially randomized---using only Python code. 
This design choice ensures that users have full control, allowing scenes to be permuted on-the-fly according to the needs of the problem being considered. 
By leveraging path tracing, photorealistic images are produced, including physically-based materials, lighting, and camera effects.
All of this can be achieved while maintaining interactive frame rates via NVIDIA's OptiX library and hardware accelerated ray tracing. %
Our tool is easily accessible via the {\tt pip} packaging system.\footnote{\tt pip install nvisii}

\begin{table}
\caption{Related work compared to our proposed system. 
`\cmark' refers to fully supported, 
`\xmark' not supported, and 
`--' partially supported or limited feature.} 
\begin{footnotesize}
\begin{tabular}{r|ccccccc}
    & AI2-Thor & iGibson & NDDS & Unity3D & Sapien & BlenderProc & Ours \\
    \hline 
    path tracing        & \xmark & \xmark & \xmark & \cmark & \cmark & \cmark & \cmark \\
    easy installation (pip)       & \cmark & \cmark & \xmark &  --     & --      & --      & \cmark \\
    cross platform          & \cmark & \cmark & \xmark & \cmark & \xmark & \cmark & \cmark \\
    Python API              & \cmark & \cmark & \xmark & \xmark & --      & --      & \cmark \\
    headless rendering      & \cmark & \cmark & \xmark & \xmark & \cmark & \cmark & \cmark \\
\end{tabular}
\end{footnotesize}
\label{tab:relatedwork}
\vspace{-4mm}
\end{table}

We offer this tool to the community to enable researchers to procedurally manage arbitrarily complex scenes for photorealistic synthetic image generation. 
Our contributions are as follows: 
1) An open source, Python-enabled ray tracer built on NVIDIA's OptiX, with a C++/CUDA backend, to advance sim-to-real and related research.
2) A demonstration of the tool's capabilities in generating synthetic images by training a DOPE pose estimator network~\citep{tremblay2018pose} and a 2D object detector~\citep{zhou2019objects} for application to real images.
3) An investigation into how physically-based material definitions can increase a pose estimator's accuracy for objects containing specular materials.

\section{Path tracer with Python interface}

\begin{figure}
    \centering
        \frame{\includegraphics[width=0.265\columnwidth]{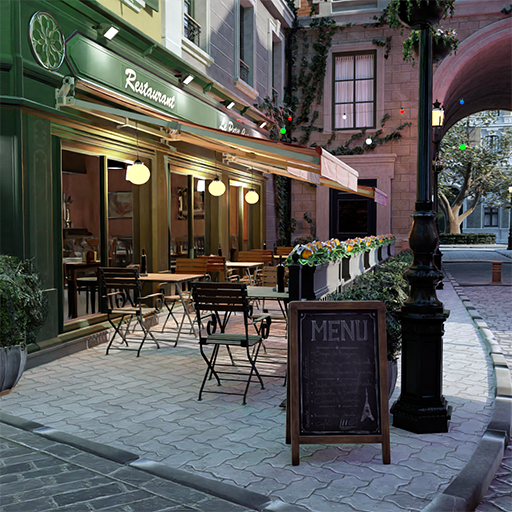}}%
        \frame{\includegraphics[width=0.265\columnwidth]{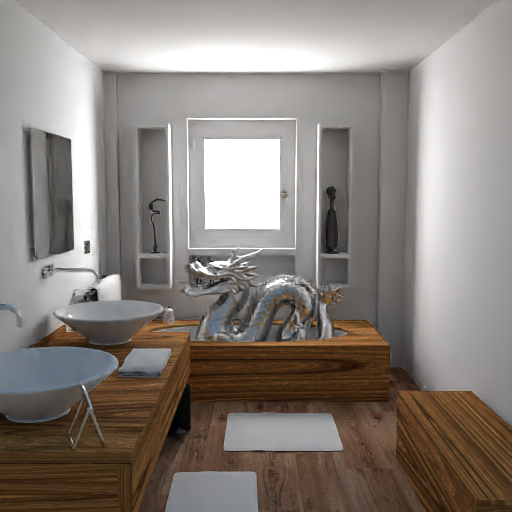}}%
        \frame{\includegraphics[width=0.265\columnwidth]{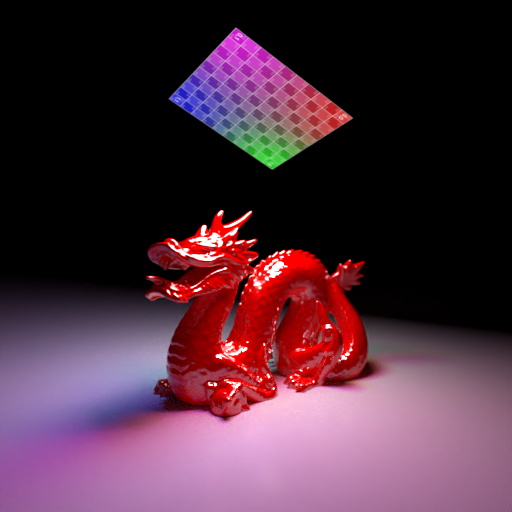}}\\
        \frame{\includegraphics[width=0.265\columnwidth]{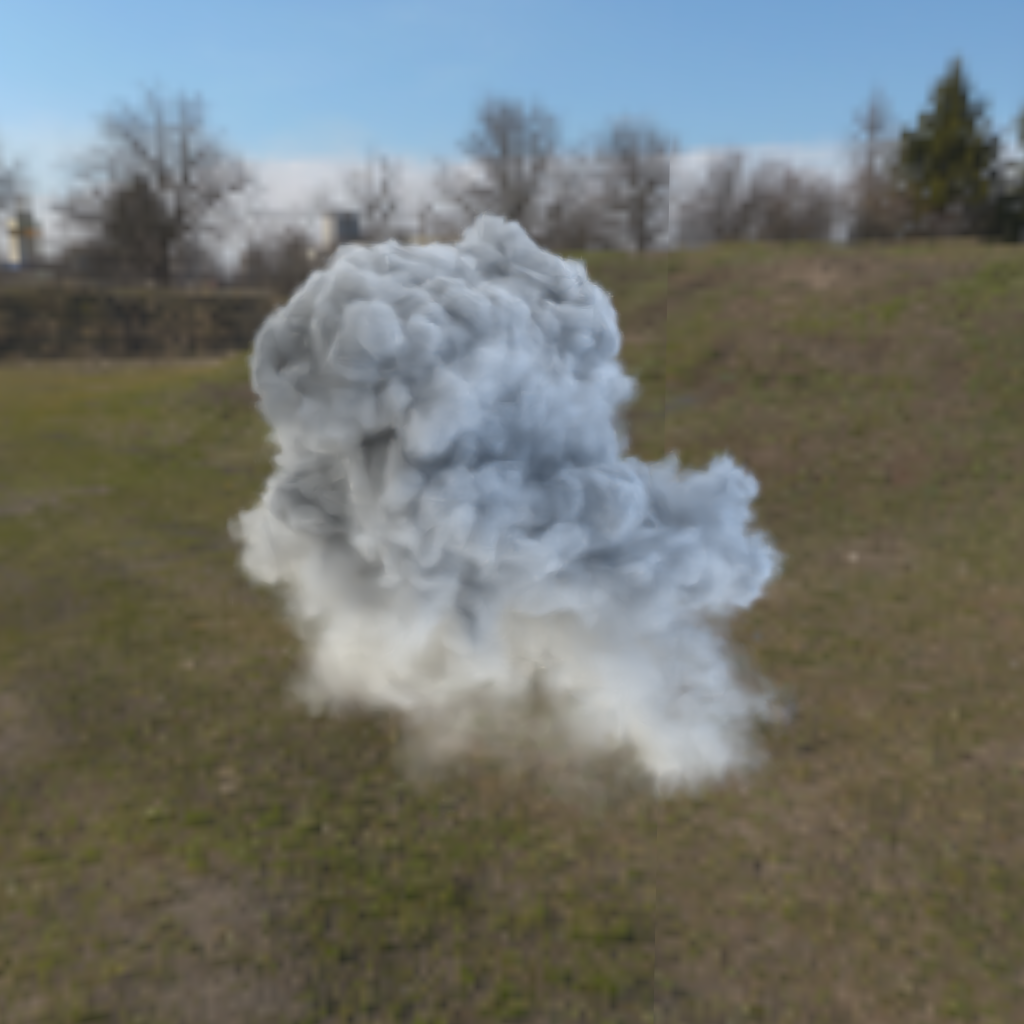}}%
        \frame{\includegraphics[width=0.265\columnwidth]{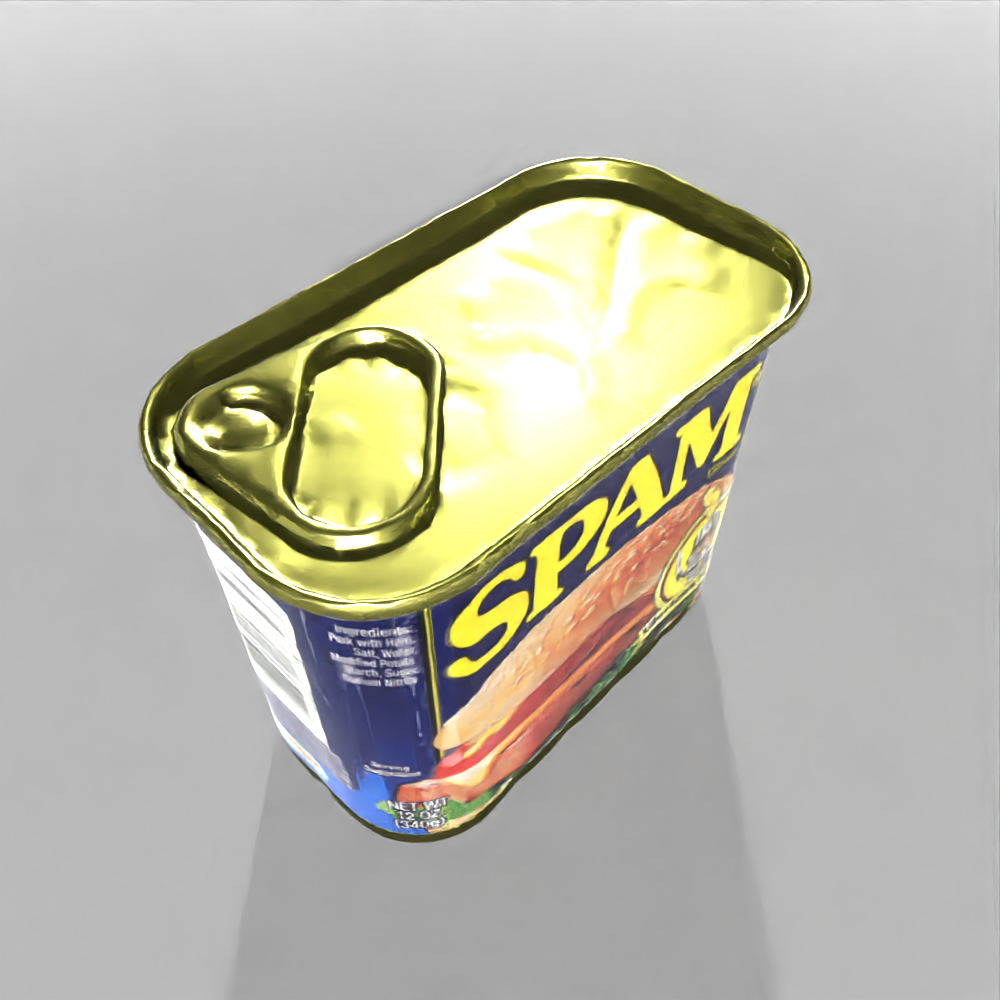}}%
        \frame{\includegraphics[width=0.265\columnwidth]{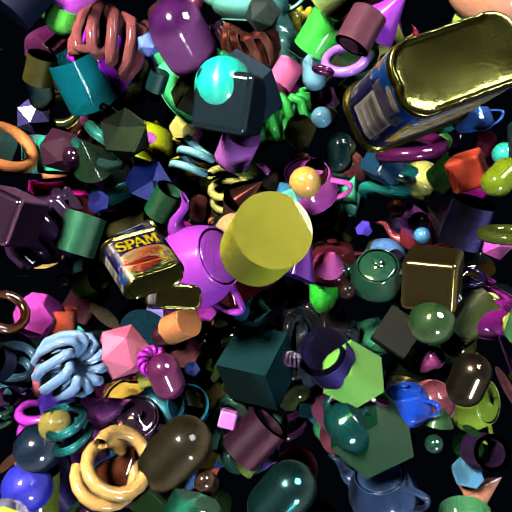}}%
    \caption{
    Example renders. 
    {\sc Top:}
    bistro scene$^{1}$
    (zoom for details),
    indoor scene$^{2}$, 
    texture used as light.
    {\sc Bottom:}
    smoke with 3D volume,
    reflective metal, 
    domain randomization. 
    }

    \label{fig:multiple}

    \vspace{-3mm}

\end{figure}

We develop the tool with three goals in mind:  
1) ease of installation, 2) speed of development, and 3) rendering capabilities. 
For ease of installation, 
we ensure that the solution is accessible, 
open source, and cross platform (Linux and Windows), 
with pre-compiled binaries that are distributed using a package manager (thus obviating the need to build the tool from source). 
For speed of development, the solution provides a comprehensive and interactive Python API for procedural scene generation and domain randomization. 
The tool does not require embedded interpreters, and it is well-documented with examples. 
Finally, we want a solution that supports advanced rendering capabilities, such as multi-GPU enabled ray tracing, physically-based materials, physically-based light controls, accurate camera models (including defocus blur and motion blur), native headless rendering, and exporting various metadata (e.g. segmentation, motion vectors, optical flow, depth, surface normals, and albedo). 
See Figure~\ref{fig:multiple} for example renders. %

\subsection{Tool architecture}

Our rendering tool follows a data-driven entity component system (ECS) design. 
Such ECS designs are commonly used for game engines and 3D design suites, as they facilitate intuitive and flexible scene descriptions by avoiding complex multiple-inheritance hierarchies required by object-oriented designs. 
This flat design allows for simpler procedural generation of scenes when compared to object-oriented designs. %

\begin{wrapfigure}{r}{0.50\textwidth}
\vspace*{-1.5em}
\begin{lstlisting}[language=Python]
  import nvisii 
  nvisii.initialize()
  # Create camera
  my_camera = nvisii.entity.create(
      name      = 'cam',
      transform = nvisii.transform.create('c_tfm'),
      camera    = nvisii.camera.create('c_cam')
  )
  my_camera.get_transform().look_at(
      eye = [3, 3, 3], at = [0, 0, 0], up = [0, 0, 1]
  )
  nvisii.set_camera_entity(my_camera)
  # Create object
  my_object = nvisii.entity.create(
      name      = 'obj', 
      transform = nvisii.transform.create('o_tfm'),
      mesh      = nvisii.mesh.create_sphere('o_mesh'),
      material  = nvisii.material.create('o_mat')
  )
  nvisii.material.get('o_mat').set_base_color([1, 0, 0])
  # Render image
  nvisii.render_to_file(
      width = 512, height = 512, 
      samples_per_pixel = 1024, 
      file_path = 'image.png'
  )
  nvisii.deinitialize()
\end{lstlisting}
\vspace*{-5.5em}
\raggedleft\includegraphics[width=1.5cm]{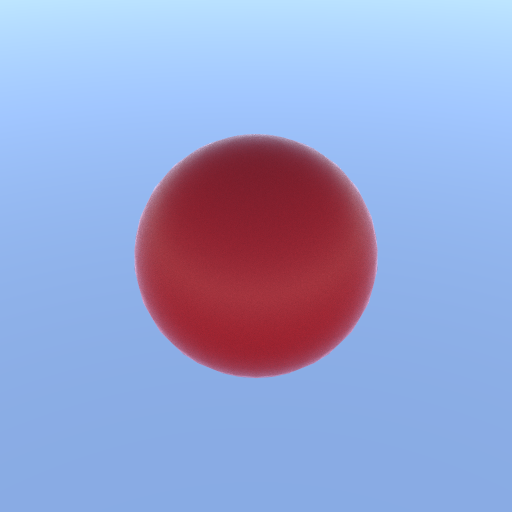}\hspace{.1em}
\caption{A minimal Python script example.}
\label{fig:sample_1}
\end{wrapfigure}

After initialization, the scene entities and components are created at runtime, as opposed to an offline scene description format like many prior solutions. 
See Figure~\ref{fig:sample_1} for a simple example.
By enabling runtime scene edits, our solution more effectively leverages modern rendering performance capabilities during synthetic image generation. 
These components can be created using either the C++ or Python API, and as these components are created and connected together using entities, the out-of-date components are asynchronously uploaded to a collection of GPUs in a data-replicated pattern for interactive rendering.

Any given entity can be attached to any combination of the following component types: transform, mesh, material, light, camera, and volume. 
Transforms refer to the entity's SE(3) behaviour, {\em i.e.,} 3D translation and rotation, which causes motion blur and so forth.
\emph{Meshes} describe the 3D points and triangles to be rendered. 
\emph{Materials} refer to the entity's 
physically-based rendering material definition.
\emph{Lights} define the energy emitted by the entity, which requires a mesh since point lights are physically impossible. 
\emph{Cameras} refer to scene views, along with focal length and 
aperture. 
\emph{Volumes} can be used in place of meshes to represent voxel data like smoke.
In addition to the aforementioned components, textures can be used to drive any material properties, and when a texture is used in concert with 
a light, the texture RGB value defines the color of the light.
Once connected to an entity, components immediately affect the appearance of the scene. 
These components can also be reused across entities as a basic form of instancing.
Swapping out and modifying components additionally serves as an effective way to randomize the scene.

\footnotetext[1]{\footnotesize{\url{https://developer.nvidia.com/orca/amazon-lumberyard-bistro}}}
\footnotetext[2]{\footnotesize{\url{https://blendswap.com/blend/12584}}}

\begin{figure}
    \centering
        \begin{tabular}{cccc}
        \includegraphics[width=0.22\columnwidth,trim=0 50 0 50,clip]{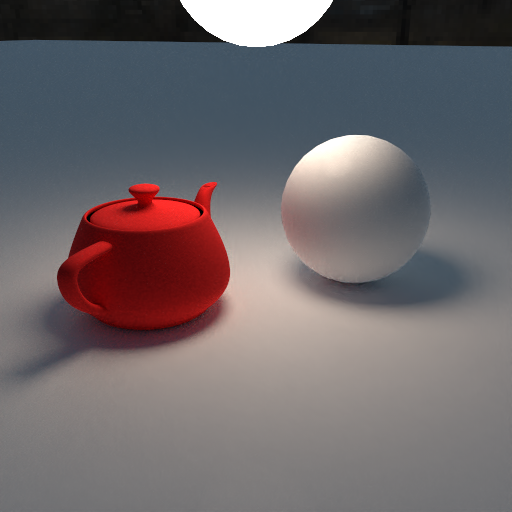} &  
        \includegraphics[width=0.22\columnwidth,trim=0 50 0 50,clip]{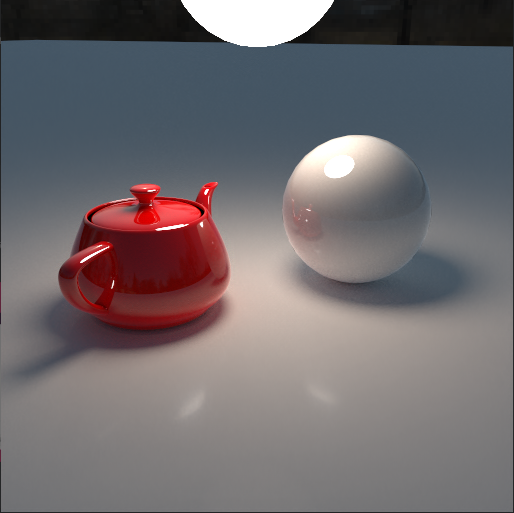} &
        \includegraphics[width=0.22\columnwidth,trim=0 50 0 50,clip]{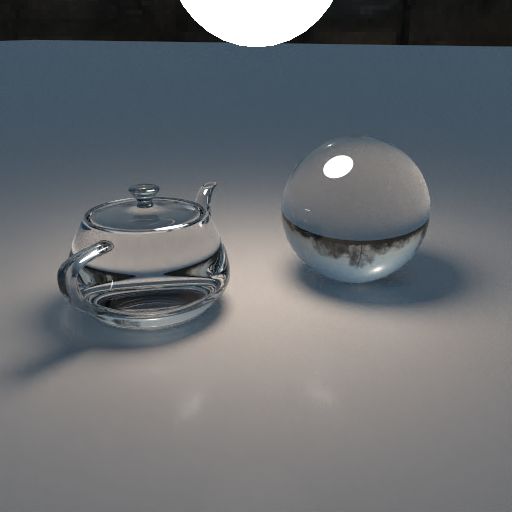} &  
        \includegraphics[width=0.22\columnwidth,trim=0 50 0 50,clip]{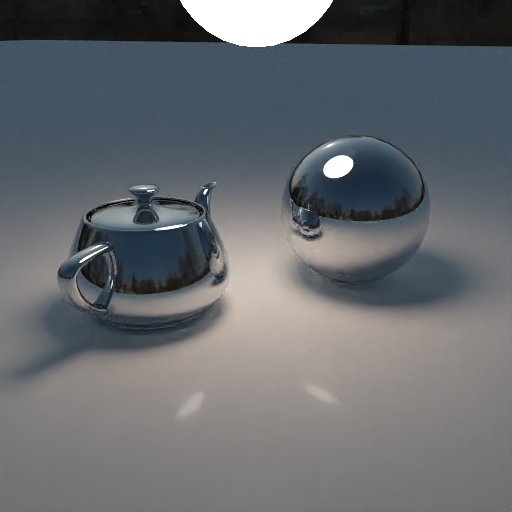}
        \end{tabular}
    \caption{Different material configurations.
    From left to right: material with roughness~=~1.0 (velvet), roughness~=~0.0 (plastic), transmission~=~1.0 (glass), metallic~=~1.0 (metal).  
    }
    \label{fig:materials}
\end{figure}

\begin{figure}
    \centering
    \includegraphics[width=0.95\columnwidth]{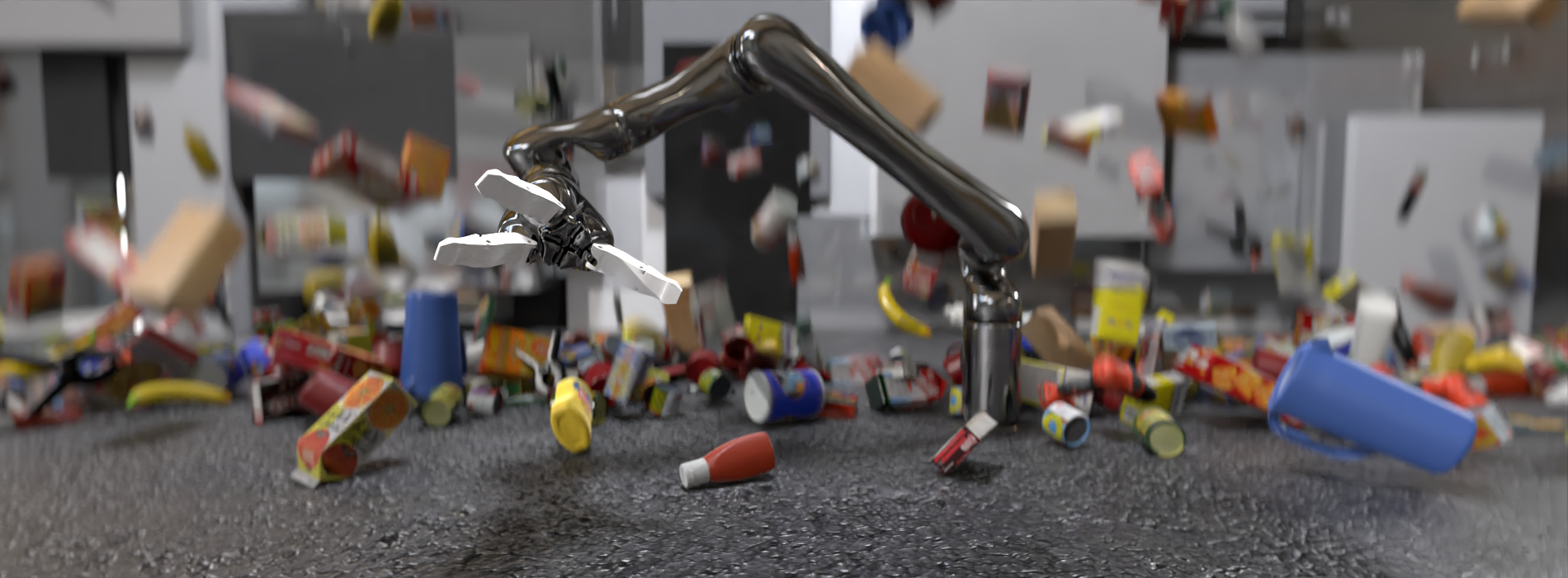}
    \caption{Motion blur and defocus blur.}
    \label{fig:defocus}      

\end{figure}

\begin{figure}
    \centering
        \begin{tabular}{cc}
        \includegraphics[width=0.45\columnwidth]{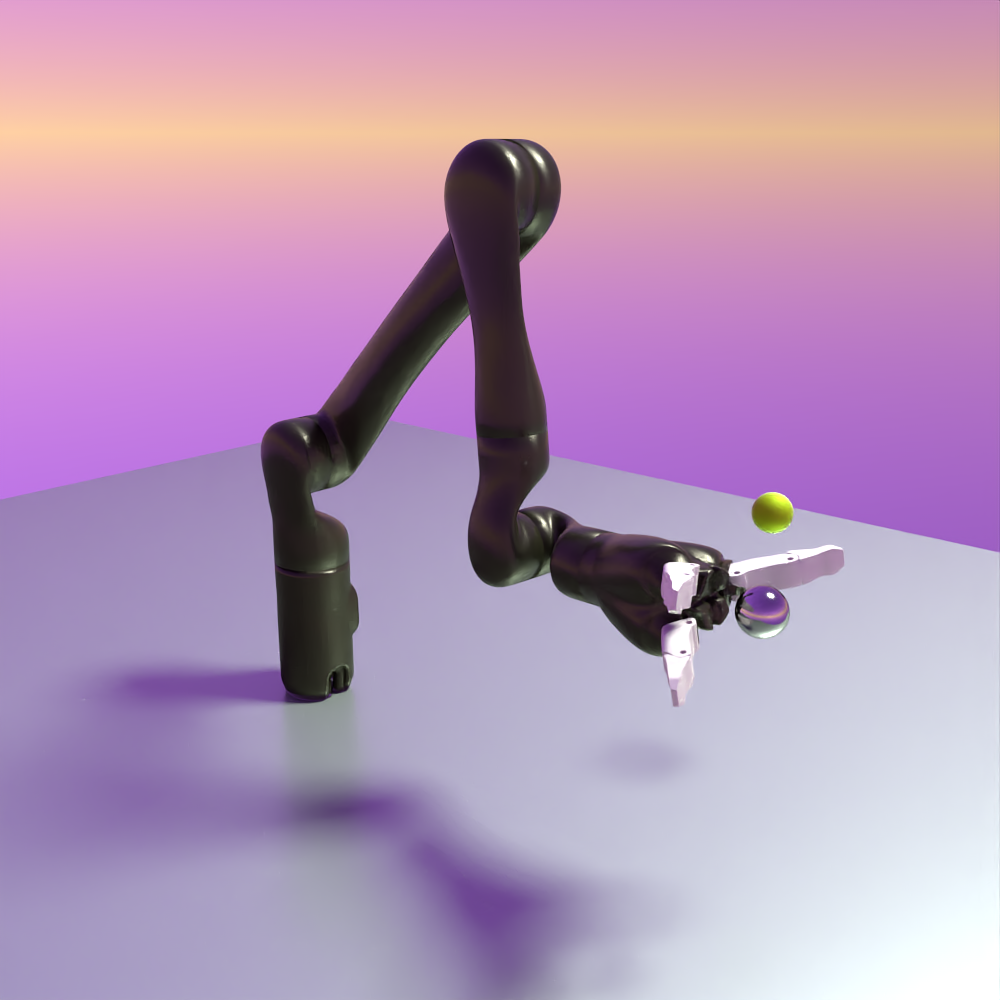} &  
        \includegraphics[width=0.45\columnwidth]{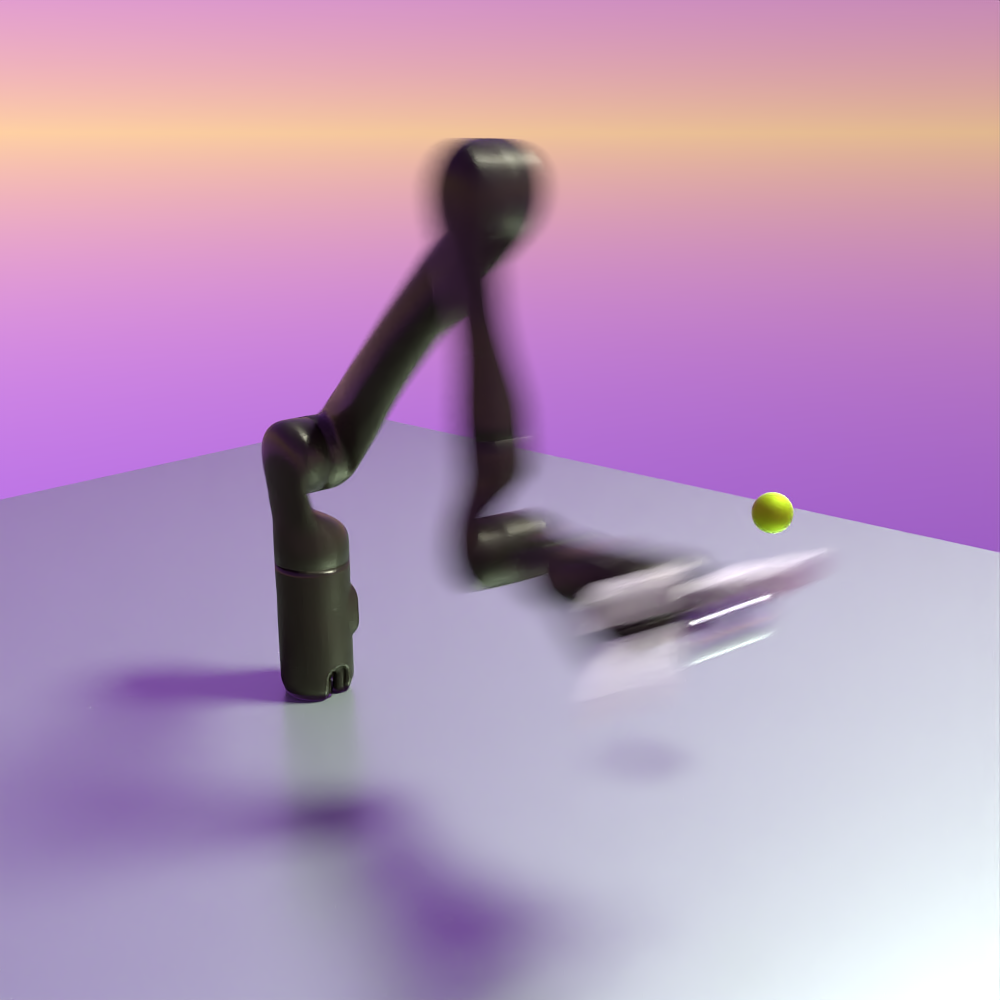}\\
        \end{tabular}
    \caption{Still image (left) and motion-blurred image (right) of a robot reaching a specific goal (green sphere).}
    \label{fig:motion_blur}
\end{figure}

\begin{figure}
    \centering
    \frame{\includegraphics[width=0.35\columnwidth]{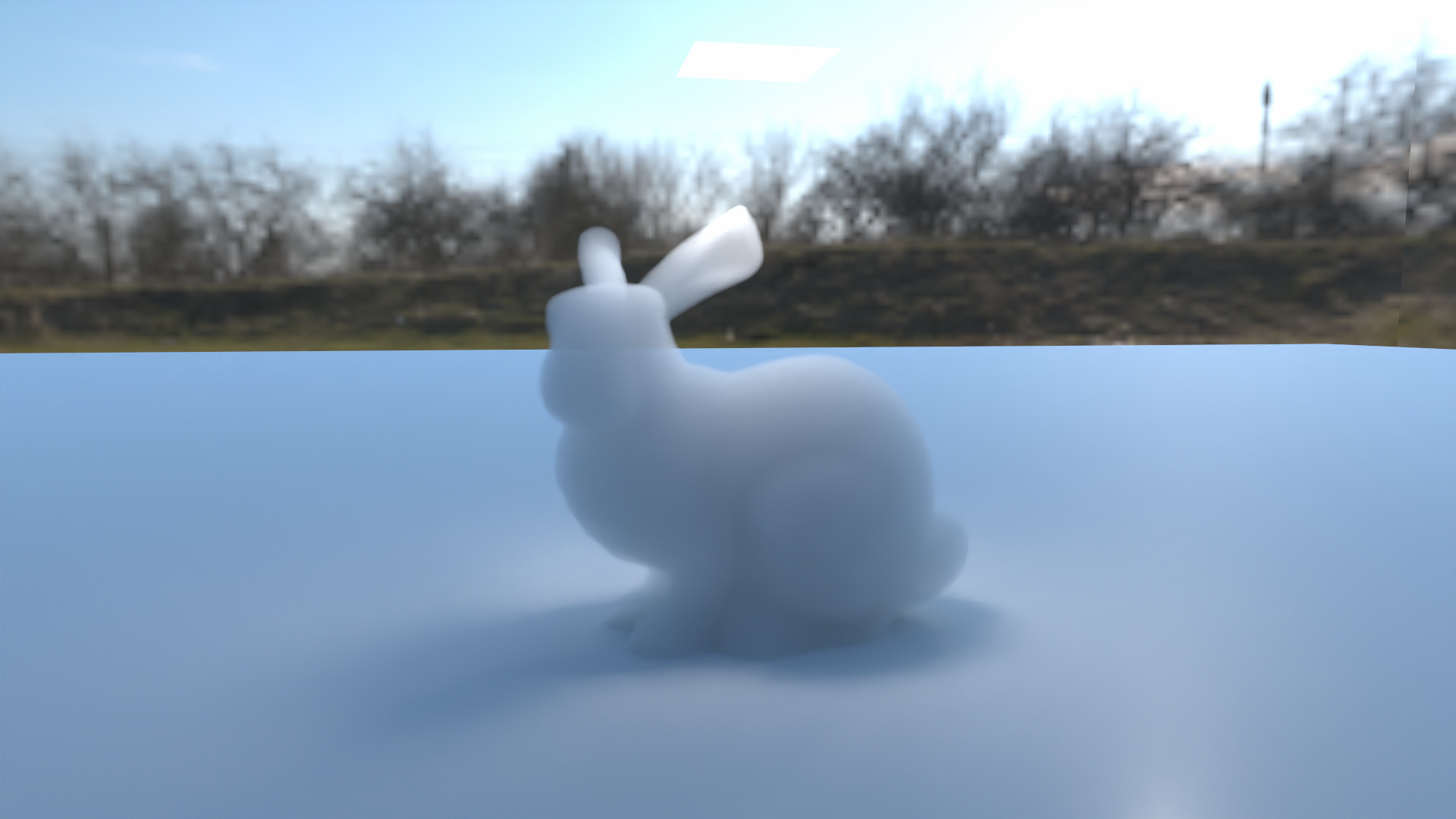}}%
    \frame{\includegraphics[width=0.35\columnwidth]{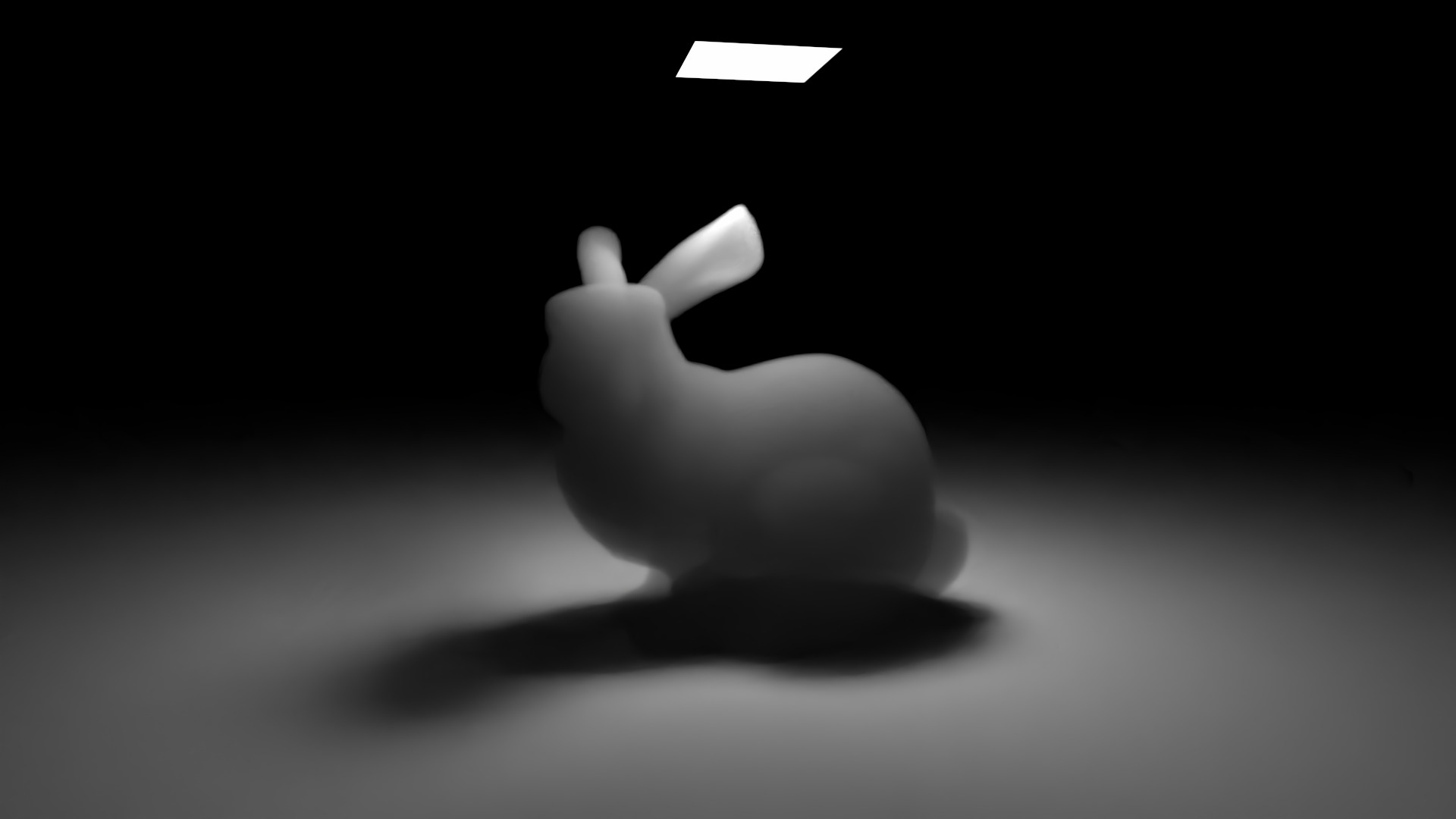}} \\
    \frame{\includegraphics[width=0.35\columnwidth]{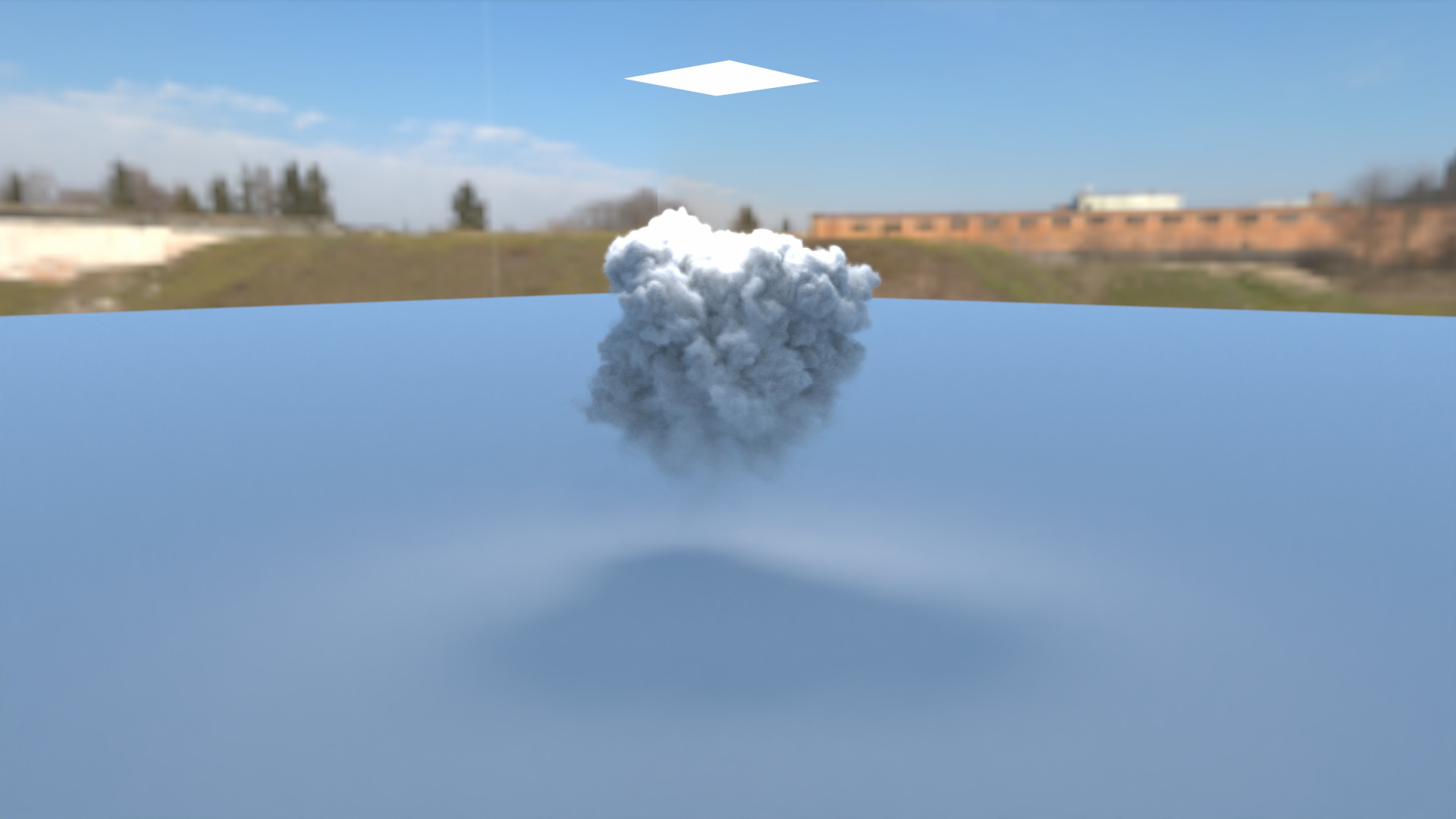}}%
    \frame{\includegraphics[width=0.35\columnwidth]{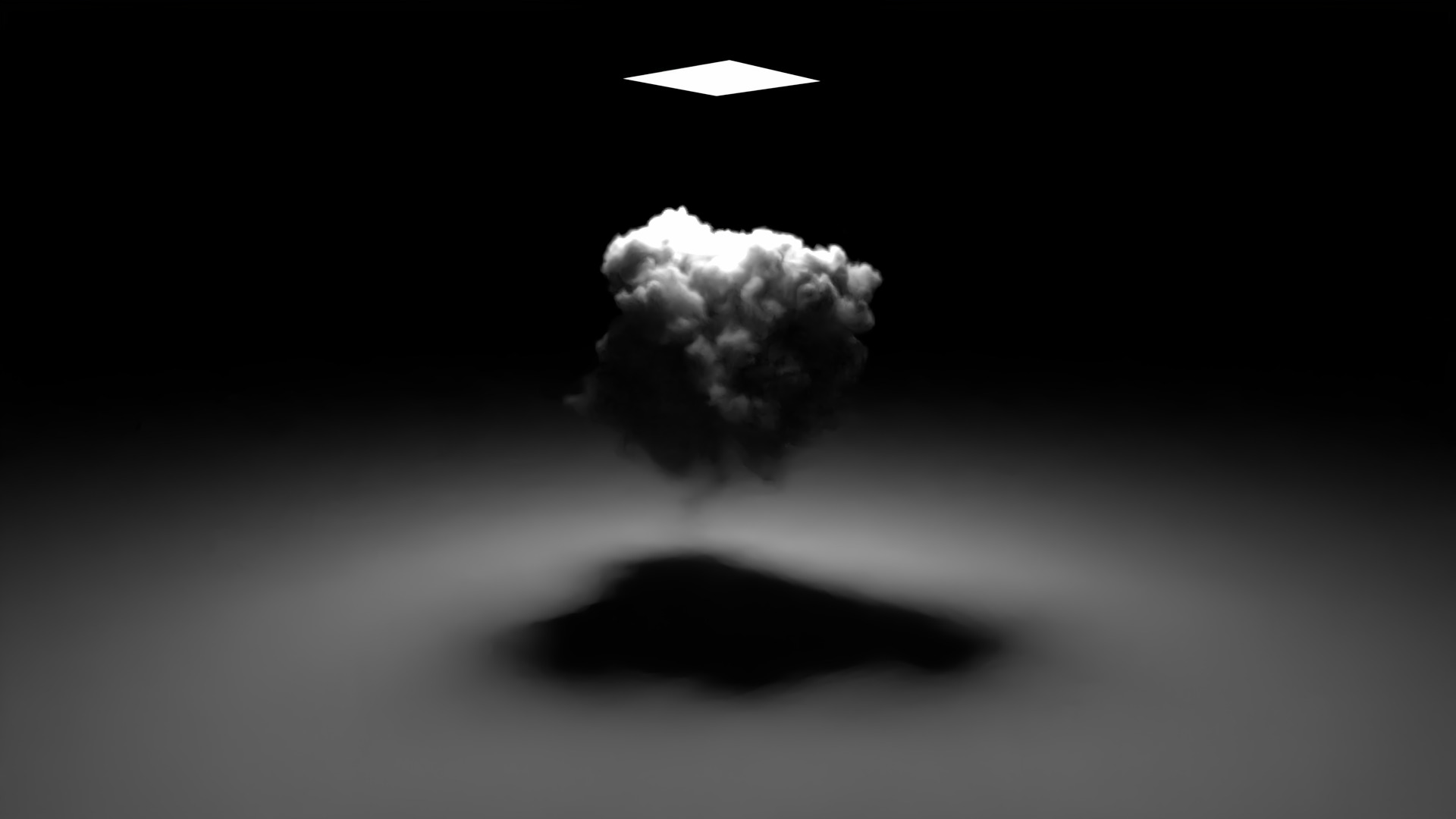}} \\
    \caption{Examples of rendering 3D volumes.}
    \label{fig:volume}
\end{figure} 

\subsection{Rendering Capabilities}

The material definition follows the Principled BSDF \citep{burley2015extending}. 
This model consists of sixteen ``principled" parameters, where all combinations result in physically-plausible, or at least well-behaved, results.
Among these are base color, metallic, transmission, and roughness parameters.
All parameters can be driven using either constant scalar values or texture components.
With these parameters, scenes can represent smooth and rough plastics, metals, and dielectric materials (\emph{e.g.}, glass), see Figure~\ref{fig:materials}.
In addition, material components also accept an optional ``normal map" texture to allow for more details on surface geometry.

Mesh components can be constructed either using a supported mesh format or alternatively by passing a list of vertices and corresponding per-vertex values. 
When meshes are created, we asynchronously construct a bottom level acceleration structure (BLAS) for later ray traversal. 
Instead of a mesh, you can also use volumetric data, such as {\em nanovdb} files or a 3D numpy array of density values, see Figure~\ref{fig:volume}.
When attaching a camera and a transform component together via an entity, users can select that entity to use as their view into the scene. 
When attaching a mesh, a material, and a transform component together via an entity, our renderer allocates an instance of the corresponding BLAS into a common instance level acceleration structure (IAS), at which point an object will appear in the scene. 
When in use, camera components can be used to induce defocus blur, and transform components can be used to induce motion blur, both of which are demonstrated in Figure~\ref{fig:defocus} and the latter in Figure~\ref{fig:motion_blur}.

Direct and indirect lighting is computed using NVIDIA's OptiX ray tracing framework~\citep{optix} through a standard path tracer with next event estimation. Multi-GPU support is enabled using the OptiX Wrapper Library (OWL) developed by \cite{wald:2020}. These frameworks enable real-world effects like reflections, refractions, and global illumination, while simultaneously benefiting from hardware-accelerated ray tracing. Finally, the OptiX denoiser is used to more quickly obtain clean images for use in training. 

\subsection{development}
An early part of development went into minimizing technical debt by automating several recurring maintenance processes like Python binding and cross-platform binary generation. 
To filter out potential OS specific compilation errors, we use continuous integration (CI) through Github actions, which at the time of writing are free for open source projects. 
We also use these CI machines as a source for binary artifacts that can be installed directly by an end user. 
For use in C++ projects, end users must still link the final binaries into their projects; however, for Python users we are able to expose the entirety of the C++ API during runtime through Python bindings that are automatically generated using SWIG. 
We also use SWIG to translate C++ Doxygen comments into Python DocStrings. 
Binary artifacts and Python bindings built by the CI service are then uploaded to the pip package manager on tagged releases for improved accessibility.
Although these design decisions are not necessarily novel, we believe these decisions make for a library that is more accessible to potential end users within the robotics and computer vision communities.

\subsection{External packaging}
Many components in our renderer expose APIs to make integration with external packages easier. 
Wherever possible, the API supports types native to the language.
For example, mesh and texture components can be constructed either from a file on disk, or alternatively from raw Python lists / NumPy arrays of vertices and pixels generated at runtime by external packages. 
We have found this to be quite useful for domain randomization, where textures can be constructed randomly using third-party pattern generator packages. 
For physics systems like PyBullet, this is also useful for constructing colliders from the vertices and indices of our mesh component. 

The API leverages existing language features for improved code legibility 
(for example by including support for keyword arguments in Python).
Camera components return either intrinsic projection matrices or affine $4 \times 4$ projection matrices, as one might be more convenient than the other depending on the domain expert using the tool. 
Finally, to help new users get started, we include comprehensive documentation through C++ Doxygen that is automatically translated to Python DocStrings during binding generation. 
Sphinx is then used to expose this documentation in an accessible website format. 
In the repository, we also include an extensive list of examples for cross-referencing.

\section{Evaluation}

In this section we explore the use of our tool to generate synthetic data for training neural networks.
We first explore the problem of 2D detections and then explore how different material propreties can be used to enhance pose estimation of metallic objects.

\subsection{HOPE object detection} 
We trained CenterNet~\citep{duan2019centernet} to perform 2D detection of known objects from the HOPE dataset~\citep{tyree2019hope} consisting of 28 toy grocery items with associated 3D models. 
Since the dataset does not contain any training images, it is an ideal candidate for sim-to-real transfer. 
The test set contains 238 unique images with 914 unique object annotations.

Using NViSII we generated 3 different dataset types: {\em DOME}, {\em MESH}, 
and {\em FAT}. 

\begin{figure}
    \centering
        \frame{\includegraphics[width=0.19\columnwidth]{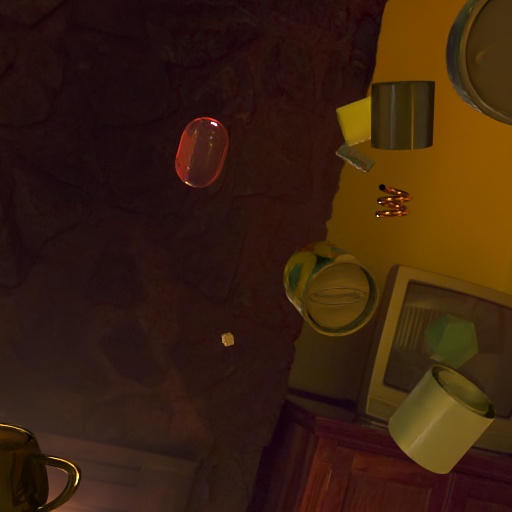}}   
        \frame{\includegraphics[width=0.19\columnwidth]{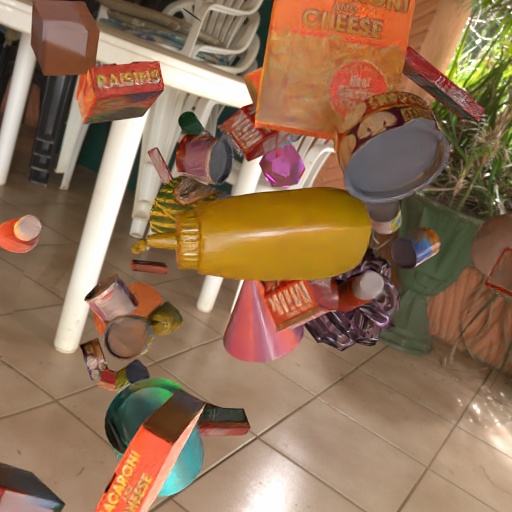}}   
        \frame{\includegraphics[width=0.19\columnwidth]{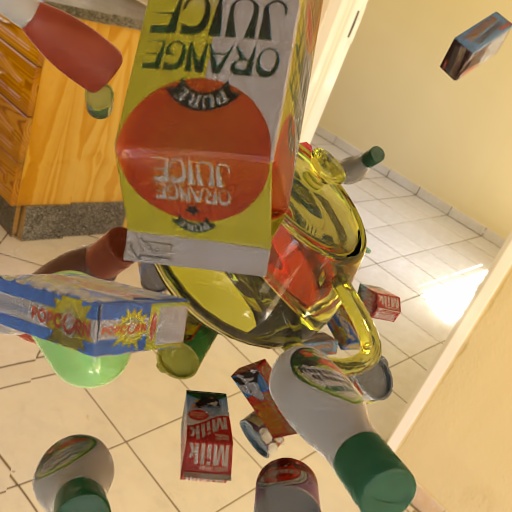}}   
        \frame{\includegraphics[width=0.19\columnwidth]{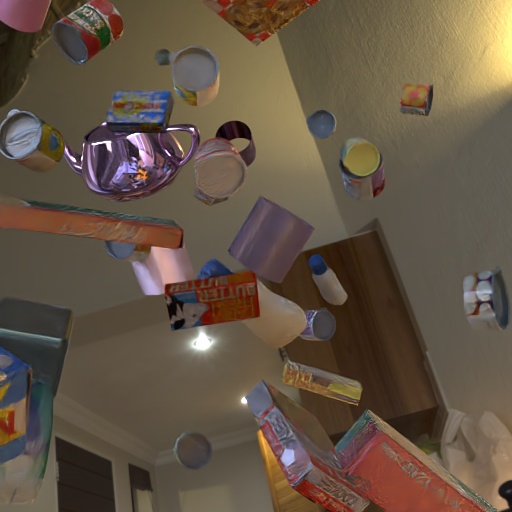}}   
        \frame{\includegraphics[width=0.19\columnwidth]{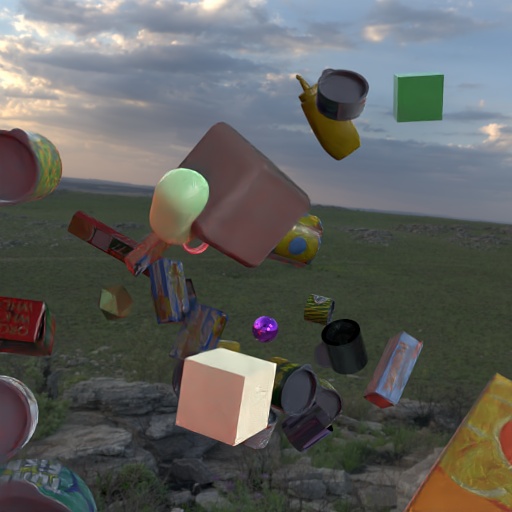}} \\
        DOME \citep{tremblay2018wad:car}
        \\
        \vspace{1mm}
        \frame{\includegraphics[width=0.19\columnwidth]{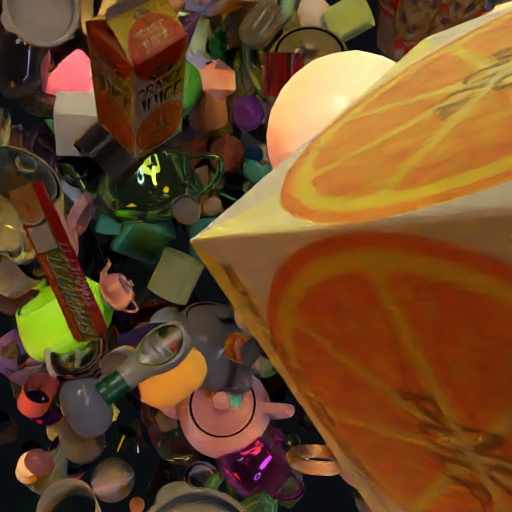}}   
        \frame{\includegraphics[width=0.19\columnwidth]{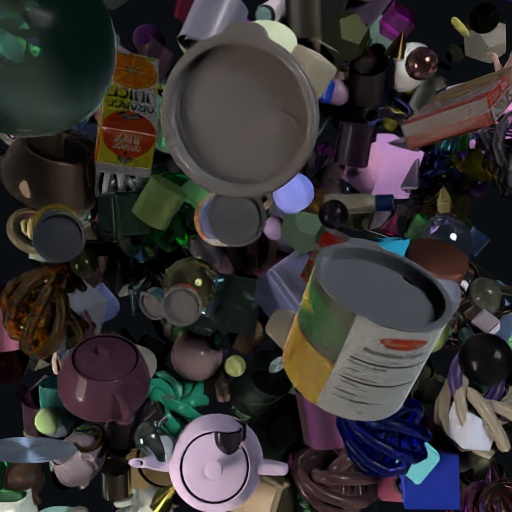}}   
        \frame{\includegraphics[width=0.19\columnwidth]{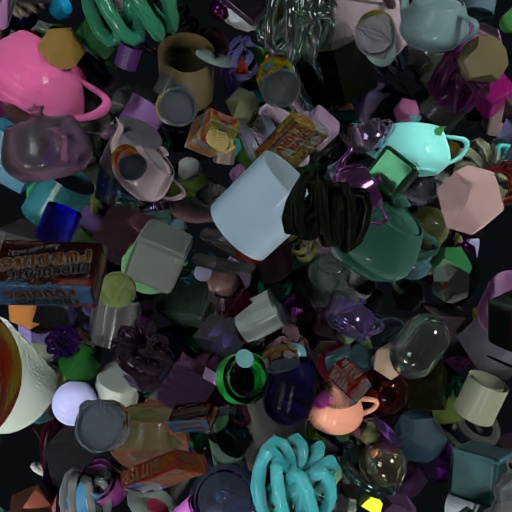}}   
        \frame{\includegraphics[width=0.19\columnwidth]{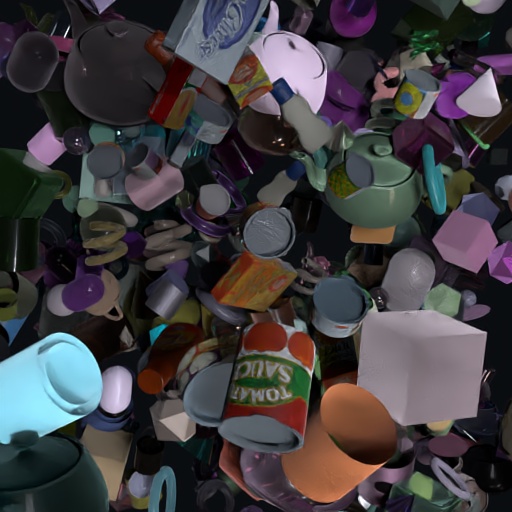}}   
        \frame{\includegraphics[width=0.19\columnwidth]{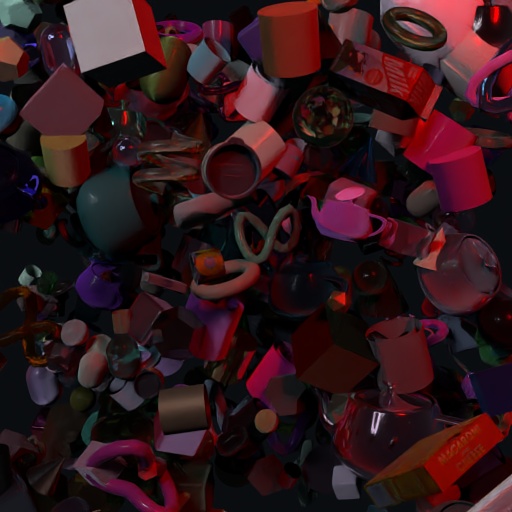}} \\
        MESH  \citep{hinterstoisser2019annotation}
        \\
        \vspace{1mm}
        \frame{\includegraphics[width=0.19\columnwidth]{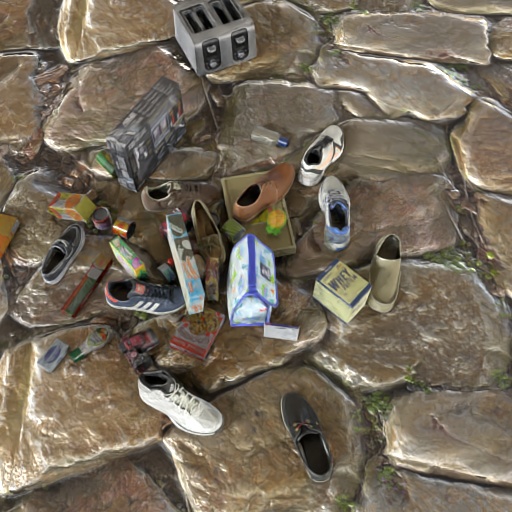}}   
        \frame{\includegraphics[width=0.19\columnwidth]{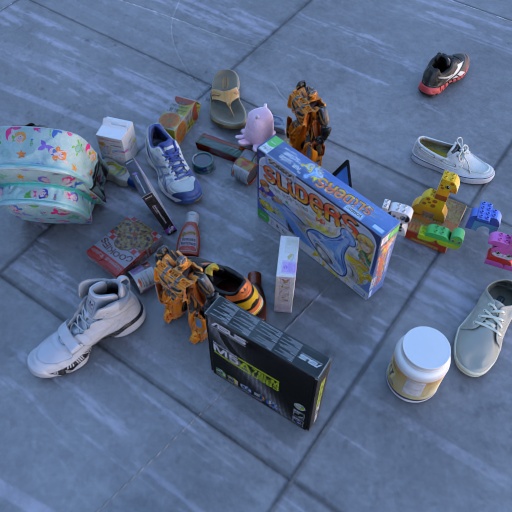}}   
        \frame{\includegraphics[width=0.19\columnwidth]{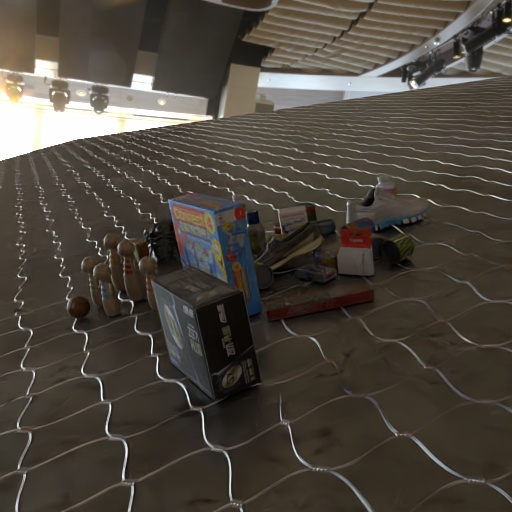}}   
        \frame{\includegraphics[width=0.19\columnwidth]{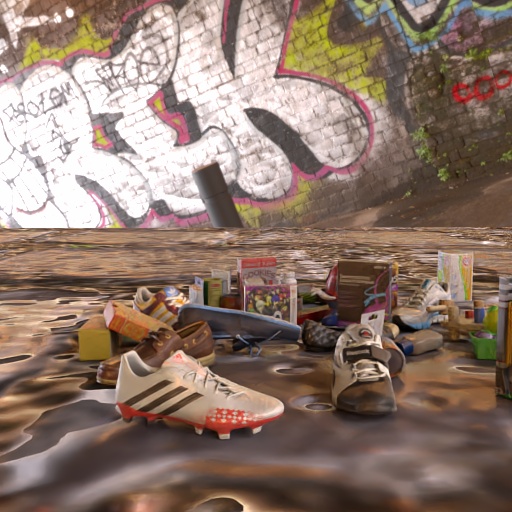}}   
        \frame{\includegraphics[width=0.19\columnwidth]{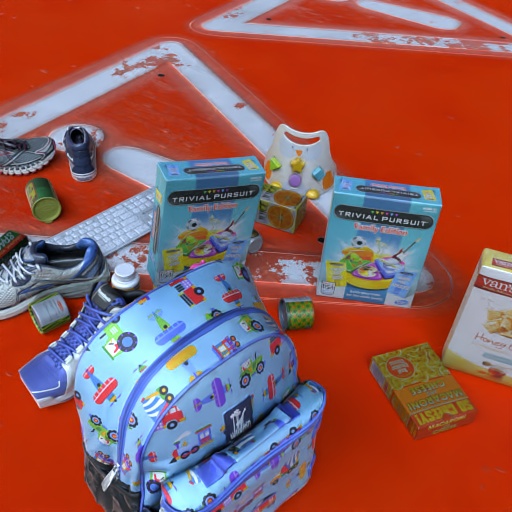}}\\         
        FAT \citep{tremblay2018arx:fat}
    \caption{
        Training images for pose estimation and object detection.
        For each row, images were generated using a procedure similar to that of the reference shown.
    }
    \label{fig:syn_train}
\end{figure}

\textbf{DOME.} \hspace{1mm} \cite{tobin2017iros:dr} demonstrated that {\em domain randomization} can be used to train a 
model fully on synthetic data to be used on real data. 
\cite{tremblay2018wad:car} extended the framework for 2D car detection. 
In this work we generate similar images,  
see Figure~\ref{fig:syn_train} first row for examples. 
In our tool, we leverage its capacity to illuminate the scene using an HDR dome texture (as in Figure~\ref{fig:multiple} top-right, with the texture spread across the upper hemisphere), which 
offers a natural looking light that mimics interreflections. 
Similar to \cite{tremblay2018wad:car} we randomize the object poses within a volume 
and add flying distractors.
See Figure~\ref{fig:syn_train} first row.

\textbf{MESH.} \hspace{1mm} \cite{hinterstoisser2019annotation} introduced 
the concept that using other 3D meshes as background could potentially 
lead to better sim-to-real. As such, we use our tool capacity to generate 
random 3D meshes and applied random material to these, we used around a 1000 moving meshes
as background as seen in Figure~\ref{fig:syn_train} second row. 
For illumination we used 2 to 6 random lights (random color and intensity) placed 
behind the camera to generate random light context. 
See Figure~\ref{fig:syn_train} second row.

\textbf{FAT.} \hspace{1mm} \cite{tremblay2018arx:fat} introduced a dataset where 
objects were allowed to freely fall into a complex 3D scene. Following on that 
work \cite{tremblay2018pose} proposed to mix falling dataset and domain randomization 
dataset to solve the sim-to-real problem. 
As such we integrated our tool with PyBullet \citep{coumans2019:pybullet} to 
let objects fall onto a simple plane, see Figure~\ref{fig:syn_train} third row. 
We simplify the prior work were we use a dome texture to create photorealistic backgrounds 
and create natural lights, and we then simply apply a random floor texture
onto the plane to simulate the surface. 
See Figure~\ref{fig:syn_train} third row.
These renders are also similar to images generated by BlenderProc \citep{denninger2019blenderproc}.

\begin{figure}
    \centering
        \frame{\includegraphics[width=0.33\columnwidth]{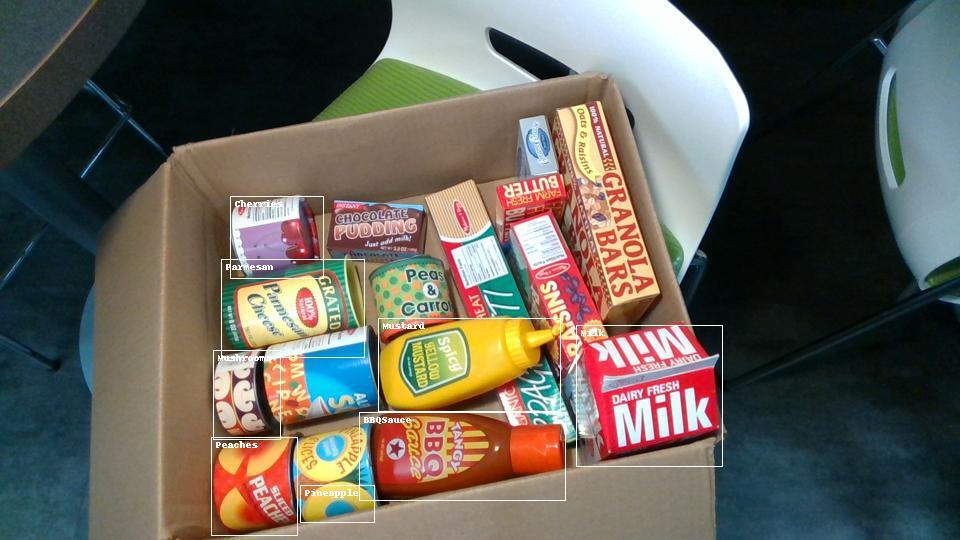}}   
        \frame{\includegraphics[width=0.33\columnwidth]{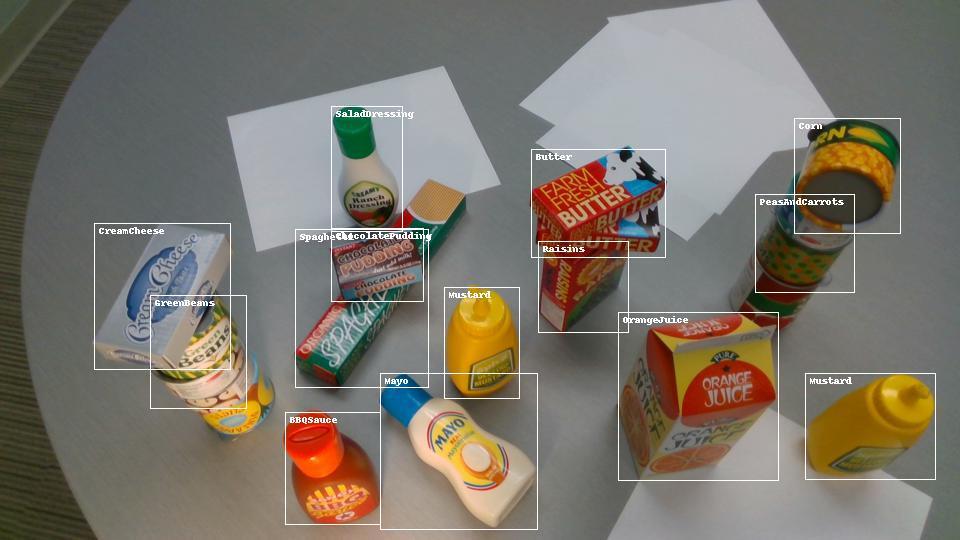}}\\   
        \frame{\includegraphics[width=0.33\columnwidth]{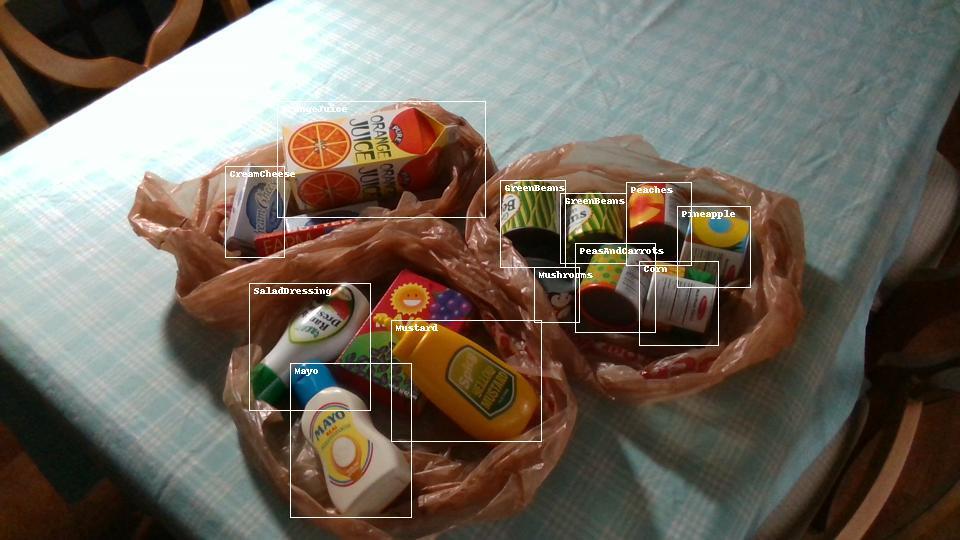}}   
        \frame{\includegraphics[width=0.33\columnwidth]{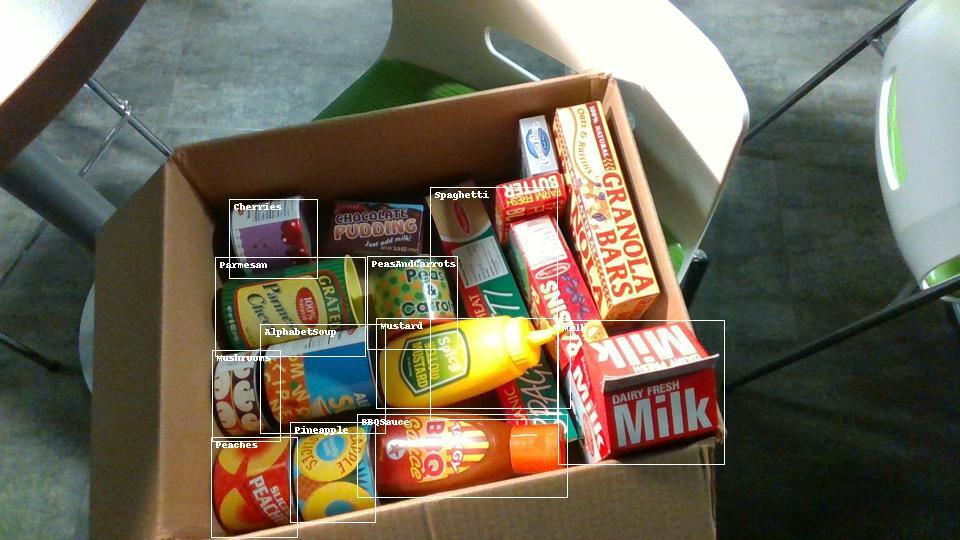}}\\   
        \frame{\includegraphics[width=0.33\columnwidth]{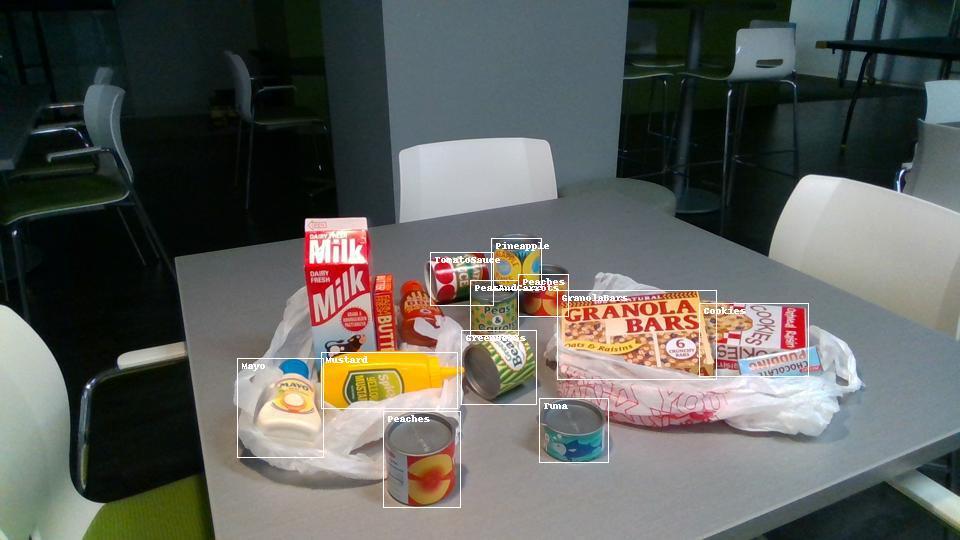}}
        \frame{\includegraphics[width=0.33\columnwidth]{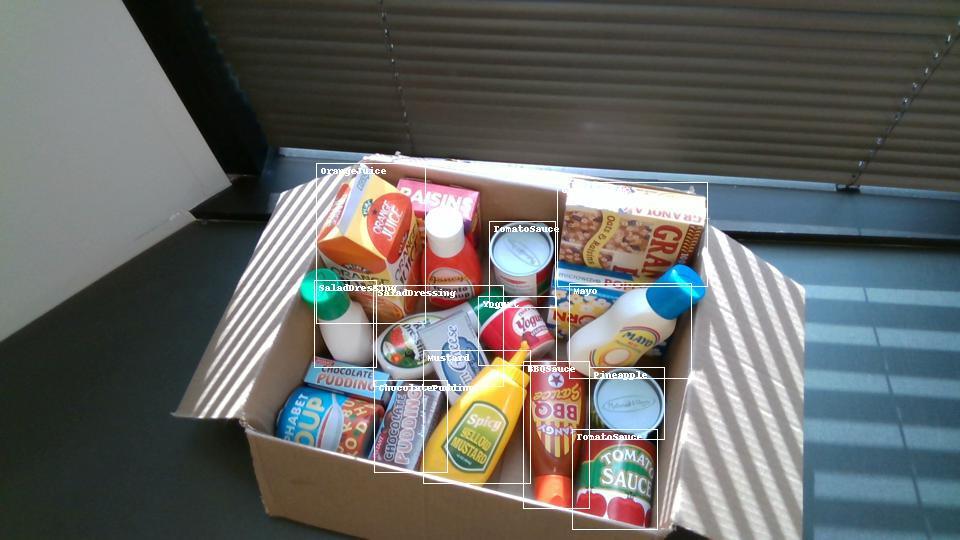}}

    \caption{
        Detections on real HOPE images~\citep{tyree2019hope} using CenterNet~\citep{duan2019centernet} trained on synthetic data generated by NViSII.
        On these images, 68.2\% of objects were detected, with no false positives.
    }
    \label{fig:detection}
\end{figure}

We use a collection of 6k rendered images to train the object detector, and we compare using a single type of dataset with a mixture of datasets, while keeping the training size constant. 
We trained CenterNet\footnote{We used the repo at \url{github.com/FateScript/CenterNet-better}}~\citep{duan2019centernet} from scratch, using SGD optimizer with learning rate of 0.02  
reduced by a factor of 10 at 81k and 108k steps, and 
a batch size of 128.  The network was trained for 126k iterations on eight NVIDIA V100s for about 12 hours. 
Table~\ref{tab:detection} shows the results, specifically that the DOME dataset performs better than MESH.
These results also confirm the observation from \cite{tremblay2018pose} that mixing 
different dataset generation methods outperfoms using a single one.  
Detections from the better model trained on using all the presented datasets 
can be seen on Figure~\ref{fig:detection}.

\begin{table}
\centering
\caption{Sim-to-real 2D bounding box detection experiment on HOPE dataset.}
\begin{footnotesize}
\begin{tabular} {r|ccc}
Methods    &   AP   &  AP50  &  AP75  \\
\hline
DOME + MESH + FAT      & \textbf{47.8} & 70.0 & \textbf{47.7}\\
DOME + MESH & 46.2 & \textbf{70.1} & 47.2\\
DOME     & 44.2 & 66.0 & 45.4\\
MESH       & 36.1 & 59.4 & 36.2\\
FAT      & 33.7 & 49.2 & 34.9\\

\end{tabular}
\end{footnotesize}
\label{tab:detection}
\end{table}

\subsection{Metallic material object pose estimator.} 

\begin{figure}
    \centering
    \frame{\includegraphics[width=0.35\columnwidth]{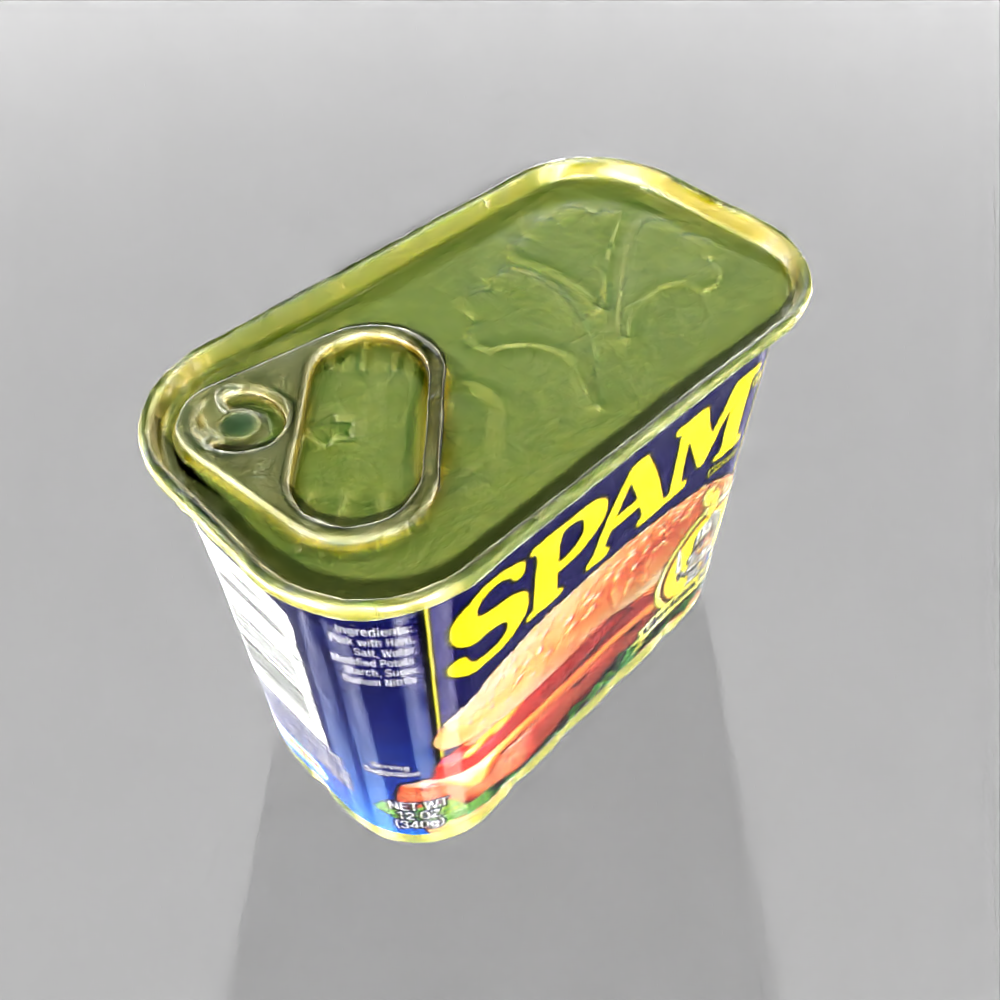}}%
    \frame{\includegraphics[width=0.35\columnwidth]{Images/meat_shiny_top.png}} \\
    \frame{\includegraphics[width=0.35\columnwidth]{Images/00000.png}}%
    \frame{\includegraphics[width=0.35\columnwidth]{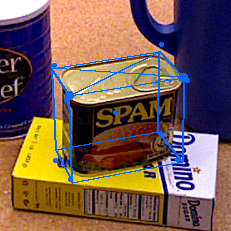}} \\
    \caption{{\sc Top-left:}  Original YCB texture with baked-in highlights.  {\sc Top-right:} Corrected flat texture and properly associated metallic material used by our tool. 
    {\sc Bottom-left:}  DR image rendered by our tool. {\sc Bottom-right:}  Pose prediction from our trained model. (Note that the model trained on the original YCB texture does not detect the meat can in this image.) 
    }
    \label{metallic}
    \vspace{-1.0em}
    \label{fig:potted_meat}
\end{figure}

In the work by \cite{tremblay2018pose}, the authors hypothesized that the somewhat disappointing pose estimation accuracy of the YCB potted meat model was caused by an inaccurate synthetic representation of the object's real-world material properties.
This model is metallic on the top and bottom, and is wrapped with a plastic-like label.
Additionally, the original 3D model provided by YCB \citep{calli2015icar:ycb} has lighting conditions (highlights) baked into the model's textures.
This results in an unrealistic appearance, especially under different lighting conditions.
To test this hypothesis, we modified the original base color texture of the model to remove all baked highlights.
Next, we manually segmented the different material properties of the object (specifically the metallic {\em vs.}~non-metallic regions) to create a more physically-accurate material description.
Figure~\ref{fig:potted_meat} compares the physically-correct material {\em vs.}~the original material with baked lighting. 
These images demonstrate how the metallic texture causes a highlight from the lights along the reflection direction under certain view angles, whereas the original texture is flat and contains unnatural highlights that do not match the surrounding synthetic scene. 
Using the approach proposed by \cite{hinterstoisser2019annotation} we generated 60k domain randomized synthetic images using path raytracer for training with random meshes and random material as background. Two to six lights were placed randomly behind the camera with randomized position, temperature, and intensity.  
We compare our results with the DOPE weights available online on the YCB-video \citep{xiang2017arx:pcnn} and we report 
the area under the ADD threshold curve. The original method scores 0.314 whereas our proposed 
solution gets 0.462.   
We see an overall performance improvement in our trained pose estimator over the same network trained with the original NDDS images. 

\section{Related Work}

Given the high demand for large annotated data sets for use in deep learning, we have seen the rise in both synthetic data sets \citep{tremblay2018arx:fat,ros2016cvpr:syn,mayer2016cvpr:flythings,handa2015arx:sn,zhang2016arx:unst,Richter_2016_ECCV,gaidon2016cvpr:vkitti} as well as a rise in tooling for generating these data sets \citep{kolve2017ai2,to2018ndds,unity2020perception,denninger2019blenderproc,xiang2020sapien}.
However, through our investigations, we have found that each of these tools come with their own set of tradeoffs when it comes to installation, rendering capabilities, and ease of development for personal solutions. 

In the field of robotics, physical simulators like PyBullet \citep{coumans2019:pybullet}, MuJoCo \citep{todorov2012mujoco}, and V-REP \citep{rohmer2013v} all have rudimentary renderers that have been used 
to generate synthetic data for robotics and computer vision applications. 
For example, Tobin {\em et al.}~\citep{tobin2017iros:dr}  introduced domain randomization, the technique of unrealistically randomizing 
3D content to bridge the reality gap, and used Mujoco's renderer to generate synthetic images.  
Similarly, domain randomization has been used to train a robot to pick up a cube by James {\em et al.}~\citep{james2017transferring}, who used 
V-REP for synthetic image generation.
PyBullet has also been used by \cite{xie2020best} to pre-train an RGBD based object segmentation method.
These physics simulators typically come with renderers that are accessible and flexible, but they lack physically based light transport simulation, material definitions, and camera effects, making them quite limited when it comes to photorealistic rendering capabilities ({\em e.g.,} no support for dielectric materials like glass, no specular materials like metals and mirrors, and no advanced camera effects like depth of field or motion blur, etc).

In order to access more advanced rendering capabilities, researchers have also explored using 
video game based renderers, such as Unreal Engine (UE4) or Unity 3D. 
For example, NDDS was introduced by To {\em et al.} as an effort to simplify the domain randomization capabilities in 
UE4 \citep{to2018ndds}.
Similarly, Qiu and Yuille added a plugin to UE4 to export meta data \citep{qiu2016arx:uncv}. 
Recently, Unity 3D introduced an official computer vision package to export metadata and 
create simple scenes for computer vision \citep{unity2020perception}.
Through these game engines, researchers have been able to explore a variety of training scenarios~\citep{long2020unrealtext,kolve2017ai2,juliani2018unity}.
Although game engines offer state-of-the-art in raster based rendering, they prioritize frame rate over image quality, and offer limited capabilities when it comes to photo-realistic material properties and light transport simulation compared to ray tracing based solutions. 
Additionally, since these applications are primarily designed for game development, they can be daunting to install, learn, modify, and script for use in synthetic data generation. 

Ray tracing based synthetic data generation has also been explored, as ray-tracing techniques more closely model the behavior of light, and can produce photorealistic images.
This exploration has mainly been done through the use of the Cycles renderer included in Blender, as Blender is an open source and easy to install solution for 3D modeling and animation.
For example, \cite{iraci2013blender} used Cycles to render transparent objects in order to train a computer vision network to detect the point cloud~\citep{sajjan2019cleargrasp}.
Blender and Cycles have also been used to generate data sets of head poses~\citep{Gu_2017_CVPR},
eye poses~\citep{wood2015_iccv}, different kind of objects~\citep{ron2020expohd}, robot poses~\citep{robotpose_etfa2019_cheind}, and so on.
In an attempt to more easily generate synthetic images, \cite{denninger2019blenderproc} introduced an extension to Blender that render falling objects onto a plane with randomized camera poses. 
These tailored solutions tend to be inflexible when users want to explore different scene configurations.
Although Blender has proven to be a useful tool for synthetic image generation, as Blender's Python API requires the use of an included, embedded interpreter, it cannot easily be scripted externally. 
As a result, solutions which use Blender for synthetic data generation tend to be constrained, and do not allow the user to easily edit and randomize the currently loaded scene. 

Other tools, like SAPIEN \citep{xiang2020sapien}, extend the Python interpreter for improved flexibility. SAPIEN optionally leverages NVIDIA's OptiX for high-quality ray tracing of articulated objects and robotics scenes; however, this functionality requires building SAPIEN from source, making their framework difficult to install. SAPIEN's material model also cannot easily be randomized and cannot easily be controlled to produce non-Lambertian materials like glass and metallics. 

\section{Conclusion}

We have presented an open-source Python-enabled ray tracer built on NVIDIA's OptiX with a C++/CUDA backend to advance sim-to-real and related research. 
The tool's design philosophy is easy install, accessible hardware requirements, 
enable scenes to be created through scripting, and  rendering of photorealistic images. 
We release NViSII in the hope that it will be helpful to the community.

\bibliographystyle{iclr2021_conference}
\bibliography{References}

\end{document}